\documentclass[final,3p,times]{elsarticle}

\usepackage{hyperref}
\usepackage{amssymb}
\usepackage{amsmath}
\usepackage{multirow}
\usepackage{placeins}
\usepackage{caption}
\usepackage{graphicx}
\usepackage{float}
\usepackage{array}
\usepackage{booktabs}
\usepackage{soul}
\usepackage{xcolor}
\usepackage[normalem]{ulem}
\usepackage[percent]{overpic}
\usepackage{makecell}
\usepackage{amssymb}
\usepackage{stfloats}   

\journal{arXiv}

\begin{document}

\begin{frontmatter}

\title{Manifold Learning for Personalized and Label-Free Detection \\ of Cardiac Arrhythmias}

\author[label1,label2]{Amir Reza Vazifeh}
\author[label1,label2,label3]{Jason W. Fleischer\corref{cor1}}

\address[label1]{Department of Electrical and Computer Engineering, 
Princeton University, Princeton, 08544, New Jersey, USA}

\address[label2]{Princeton Precision Health, Princeton University, 
Princeton, 08544, New Jersey, USA}

\address[label3]{Omenn-Darling Bioengineering Institute, Princeton University, 35 Ivy Lane, Princeton, 08540, New Jersey, USA}

\cortext[cor1]{Corresponding author: jasonf@princeton.edu}

\begin{abstract}
\textcolor{black}{Electrocardiograms (ECGs) provide non-invasive measurements of heart activity and are well-established tools for detecting cardiac arrhythmias. Although supervised machine learning has emerged as a promising approach for automated heartbeat classification, substantial variations in ECG signals across individuals and leads, combined with inconsistent labeling standards and dataset biases, make it difficult to develop generalizable models. Dimensionality reduction is an unsupervised approach that maps high-dimensional data into a lower-dimensional space while preserving the underlying structure, enabling data visualization and pattern discovery. Conventional methods, e.g., principal component analysis, prioritize large (and often obvious) variances in the data and typically overlook subtle yet clinically relevant patterns. Here, we show that nonlinear dimensionality reduction (NLDR) algorithms, e.g., t-distributed stochastic neighbor embedding (t-SNE) and uniform manifold approximation and projection (UMAP), can identify medically relevant features in ECG signals without pretraining or prior information. Using the MIT-BIH Arrhythmia Database, we show that: a) applying NLDR to a mixed population of heartbeats reveals inter-individual differences in morphology, as signals from the same person cluster together in latent spaces; and b) applying NLDR to the heartbeats of a single individual separates normal beats from arrhythmias into distinct clusters, which can be identified by clustering algorithms in an unsupervised manner. To our knowledge, this is the first systematic evaluation of NLDR for unsupervised arrhythmia detection. As a proof-of-concept, we demonstrate that both UMAP and t-SNE achieved trustworthiness scores of $\geq0.95$, indicating that local neighborhoods from the original space are well preserved in the embedding. Moreover, classification on the 2D embeddings outperforms classification in the original high-dimensional space, with a simple k-NN classifier discriminating individual recordings with $\geq80\%$ accuracy and identifying arrhythmias with a median accuracy of $\geq98\%$ and median F1-score of $\geq 85\%$. These results show that NLDR holds much promise for cardiac monitoring, and for personalized healthcare beyond cardiology.}
\end{abstract}


\end{frontmatter}

\section{Introduction}
\label{Introduction}
\begin{figure}[htbp]
    \centering
\includegraphics[width=\columnwidth]{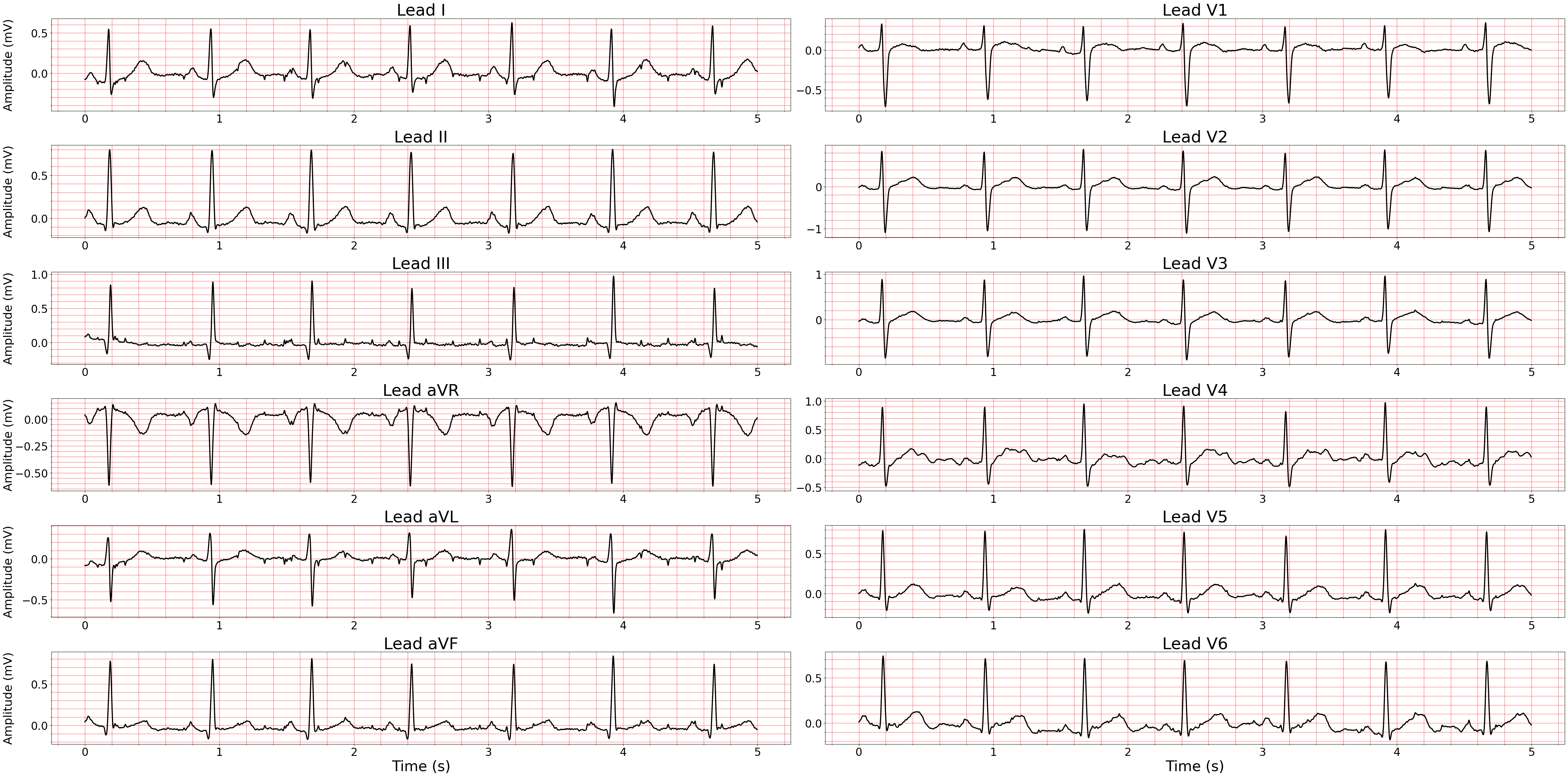} 
    \caption{ECG signals from multiple leads in a standard 12-lead setup. The data is from recording A6371 of the CPSC2018 dataset \cite{liu_open_2018}.}
    \label{fig:12leadecg}
\end{figure}

Cardiovascular disease is the leading cause of death globally~\cite{di_cesare_heart_2024}. The United States is no exception, with the Centers for Disease Control and Prevention reporting 702,880 deaths attributed to heart disease in 2022 alone~\cite{carbone_smoking_2025}. Among them, cardiac arrhythmias, which affect an estimated 1.5\% to 5\% of the general population, are associated with significant morbidity and mortality. The symptoms of cardiac arrhythmias can vary widely or be absent altogether, and because they occur unpredictably, it is difficult to estimate how common they are \cite{Desai2023Arrhythmias}. An essential tool for detecting arrhythmias is the electrocardiogram, which is a non-invasive diagnostic method that offers critical information about the heart's electrical activity, rhythm, and overall health~\cite{kligfield_recommendations_2007}. A 12-lead ECG is the standard setup for capturing electrical signals from the heart; in this configuration, 10 electrodes are placed on the body (four on the limbs and six on the chest) to record nerve depolarization and repolarization from 12 different angles. This results in 12 different waveforms for each heartbeat, labelled according to three bipolar limb leads (I, II, III), three augmented unipolar limb leads (aVR, aVL, aVF), and six precordial leads (V1–V6). However, not all of these signals are independent, e.g., the signal from aVF can be derived from aVL and aVR. Among the bipolar leads, lead II is often preferred because it is most aligned with the heart's major axis and gives the strongest measurement of pumping from the left ventricle~\cite{rajoub_machine_2020}. Figure \ref{fig:12leadecg} illustrates an example of signals from different leads in a 12-lead ECG setup.

Manual analysis of ECGs is time-consuming, subjective, and prone to errors \cite{cook_accuracy_2020, breen_ecg_2022}. As a result, there has been a growing interest in machine learning approaches to enhance the accuracy and efficiency of heart arrhythmia detection using ECGs. Conventional methods extract user-defined features from ECG signals (such as statistical, Fourier, or morphology-based features), followed by the use of various classifiers (e.g., Support Vector Machines or K-nearest neighbors) \cite{usha_kumari_automated_2021, baraeinejad_design_2022, ebrahimzadeh_detection_2010, epmoghaddam_graph-based_2024, bertsimas_machine_2021, sharma_wavelet_2019, AlShammary-Leibler_2024}, or cluster analysis techniques \cite{yeh_analyzing_2012}.
More recently, deep learning has gained prominence for its ability to automatically extract features and classify heartbeats \cite{yu_automatic_2021, lai_non-standardized_2020, singh_novel_2024, anand_enhanced_2024, rajkumar_arrhythmia_2019, daydulo_cardiac_2023, chen_implementing_2024, cui_deep_2023, akkus_spindle_2025}. Although deep learning-based approaches often outperform traditional methods in accuracy, their training procedures are brittle with respect to initial conditions and dataset breadth, and their operation and results lack interpretability.

\begin{figure*}[htbp]
    \centering
    \includegraphics[width=\textwidth]{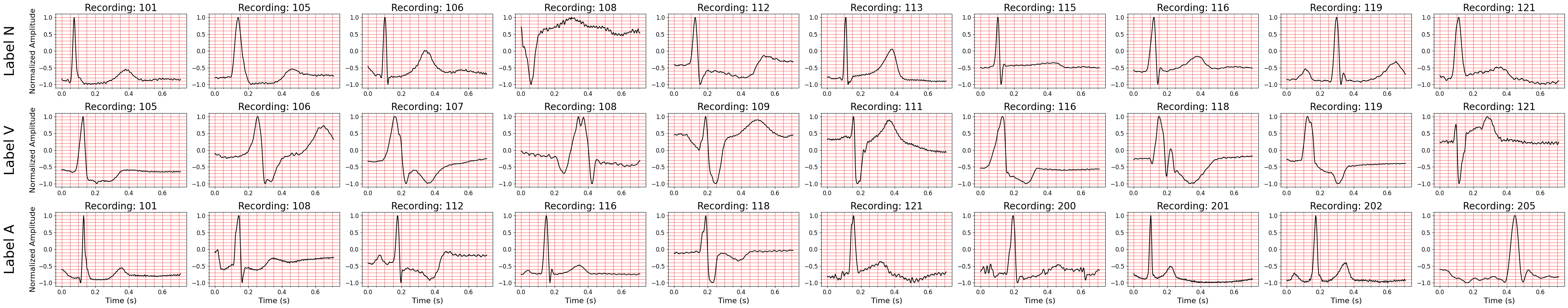} 
    \caption{Different heartbeat profiles. Each subplot is sourced from different individuals in the MIT-BIH dataset, all recorded using the modified limb lead II (MLII). Top row: Normal beats (N), Middle row: 
    Premature ventricular contraction (V), Bottom row: Atrial premature beat (A).}
    \label{fig:different_people_mit_bih}
\end{figure*}

Most methods of automated heartbeat analysis still rely on supervised learning, which requires large labeled datasets. However, many factors contribute to the variability in ECG signals, making it challenging to design a generalizable model. First, standard 12-lead ECG systems record 12 distinct waveforms for each heartbeat (Figure \ref{fig:12leadecg}), and separate models must be trained for each lead. Further, not all datasets include signals from all leads. For example, the MIT-BIH dataset \cite{moody_impact_2001, goldberger_physiobank_2000}, widely regarded as the benchmark for arrhythmia detection, provides only two leads per recording: every patient has data from lead II, but the other lead varies between patients (either V1, V2, V4, or V5). Moreover, if electrodes are connected in slightly different positions during the test phase (i.e., after the model has been trained), then the signal quality may change, leading to potential model failure.  Another issue is that signals with the same class label but from different individuals exhibit strong variability (Figure \ref{fig:different_people_mit_bih}). These individual variations can reduce the model's robustness to new, unseen data, further complicating efforts to train a generalizable model. In addition, training datasets often exhibit bias, both in terms of demographic representation and arrhythmia types. For example, in the MIT-BIH dataset, over 80\% of the signals correspond to normal beats (Figure \ref{fig:aami_distribution}). Among the arrhythmias, the majority are premature ventricular contractions (PVCs). One consequence is that many studies have focused on designing machine learning models to detect specific heart patterns, especially PVC \cite{yu_automatic_2021, sarshar_premature_2022, de_marco_classification_2020, mastoi_machine_2021, jun_premature_2016}, making them unable to classify other types of heart arrhythmias. Moreover, there are different standards for labeling ECG signals \cite{merdjanovska_framework_2023}, and supervised learning can only predict labels present in the training data.

\begin{table*}[t]
\centering
\begin{minipage}{0.45\textwidth}
    \centering
    \scriptsize
    \renewcommand{\arraystretch}{1.2}
    \captionof{table}{\textcolor{black}{Overview of ECG dataset}}
    \label{tab:dataset_overview}
    \begin{tabular}{c|c}
    \hline
    \bf Parameter & \bf MIT-BIH \cite{moody_impact_2001} \\
    \hline
    Number of Recordings & \begin{tabular}[c]{@{}c@{}}48 \\ (Our Study: 40)\textsuperscript{*}\end{tabular} \\
    \hline
    Recording Length & $\sim$ 30 mins \\
    \hline
    Number of Leads & \begin{tabular}[c]{@{}c@{}}2 \\ (mostly MLII, V1)\end{tabular} \\
    \hline
    Sampling Frequency & 360 Hz \\
    \hline
    R-peak Location & Annotated \\
    \hline
    Label for each heartbeat & Yes \\
    \hline
    Number of Heartbeats & \begin{tabular}[c]{@{}c@{}}112,551 \\ (Our study: 97,117)\textsuperscript{*}\end{tabular} \\
    \hline
    \end{tabular}

    \vspace{0.3cm}
    \parbox{0.9\linewidth}{
        \scriptsize\textsuperscript{*} For consistency, we selected MIT-BIH recordings with MLII as the first lead and V1 as the second. This reduced the dataset to 40 recordings and 97,117 heartbeats.
    }
\end{minipage}
\hfill
\begin{minipage}{0.45\textwidth}
    \centering
    \includegraphics[width=\linewidth]{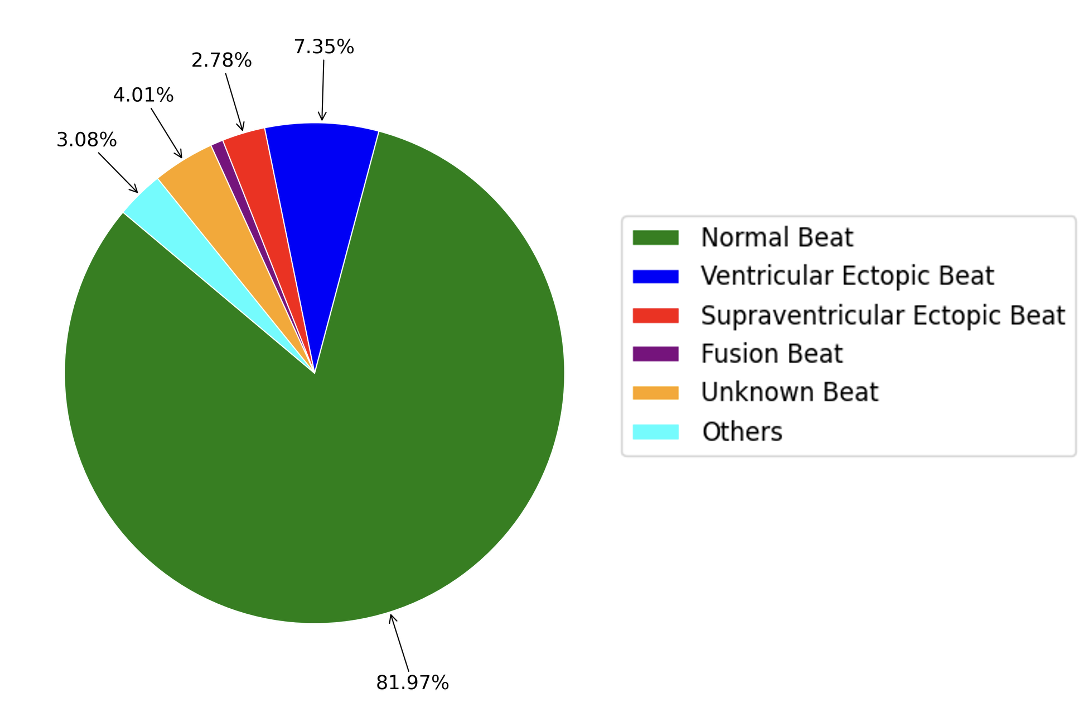}
    \captionof{figure}{Class distribution of AAMI labels in the MIT-BIH dataset.}
    \label{fig:aami_distribution}
\end{minipage}
\end{table*}

Given these challenges, there is a growing need to explore unsupervised techniques that do not require labeled data and can handle the inherent variability in ECG signals. A promising approach is manifold learning, a method based on the hypothesis that high-dimensional data in the real world often lies on low-dimensional latent manifolds within the high-dimensional space \cite{fefferman_testing_2016}. More specifically, dimensionality reduction algorithms try to improve data visualization by transforming the original high-dimensional data space \( X \in \mathbb{R}^{N \times n} \) (containing \( N \) samples, each with \( n \) dimensions), into a lower-dimensional space with \( d \) dimensions (where \( d \ll n \)). The paradigm example is PCA, which is a linear method that reduces dimensionality in two stages: 1) rotating the data to align with the directions of variance, and 2) keeping the $d$ largest directions (typically two). More recent methods extend the analysis to other data properties, such as the distance and topology between neighboring points. These algorithms are inherently nonlinear and try to preserve as much of the original neighborhood and graph structure as possible.

The two most common neighbor-embedding methods are t-SNE and UMAP. The former fits points in the original data space with a Gaussian distribution and points in the latent space with a Cauchy distribution (the fatter tails prevent widely separated points in the original space from being crowded together as the dimension is reduced). The latter extends this technique in two ways: 1) using many Gaussians in the original data space to adjust for differences in local point density and 2) adding two extra parameters in the target Cauchy distribution to adjust the width and fall-off of the tails. Compared to t-SNE, UMAP offers superior runtime performance, better preservation of global structure, and the ability to scale to significantly larger datasets \cite{JMLR:v9:vandermaaten08a, mcinnes_umap_2018}. Both algorithms have been extensively used in biomedical research, e.g. detecting rare cells in flow cytometry \cite{weijler_detecting_2021, van_unen_visual_2017} and RNA sequencing \cite{becht_dimensionality_2019}, identifying prognostic tumor subpopulations \cite{abdelmoula_data-driven_2016} and individuals with Parkinson’s Disease \cite{oliveira_use_2018}, phenotyping microglia cells \cite{colombo_tool_2022} and COVID-19 patients \cite{fleischer_late_2020}, and detecting outliers in large radiological datasets \cite{islam_outlier_2025}.

In this study, we apply dimensionality reduction algorithms to heartbeats extracted from the MIT-BIH dataset. We first consider all heartbeat waveforms from all subjects using the available leads and show that the latter methods isolate most patients into individual clusters. We then re-apply the algorithms to each subject, revealing heartbeat variations for individual patients. Our key contributions include:

\begin{itemize}
    \item The introduction of unsupervised NLDR methods to visualize high-dimensional heartbeat waveforms in lower-dimensional latent spaces. These algorithms can sort data into recognizable groups, without pre-applying labels, and can operate on ECG signals from different leads, without modification.

    \item The demonstration that signals from different patients are distinctly separated in the 2D latent space. This suggests that heartbeat signals are highly individual-specific and can serve as a unique fingerprint. Consequently, relying on signals from different individuals to train a supervised model may not be effective for heart arrhythmia detection in new individuals.
    
    \item The demonstration that heart arrhythmias can be detected in an unsupervised, personalized, and lead-independent manner by applying NLDR to signals from each individual separately. \textcolor{black}{This approach produces distinct clusters in low-dimensional spaces, corresponding to different heartbeat phenotypes, that distinguish normal beats from arrhythmias and can be identified by clustering algorithms.} It is not limited to specific classes and can handle multiple types of cardiac anomalies and subtle ECG signal variations.
\end{itemize}
\section{Methodology}
\label{method}

\begin{figure*}[htbp]
    \centering
    \includegraphics[width=\textwidth]{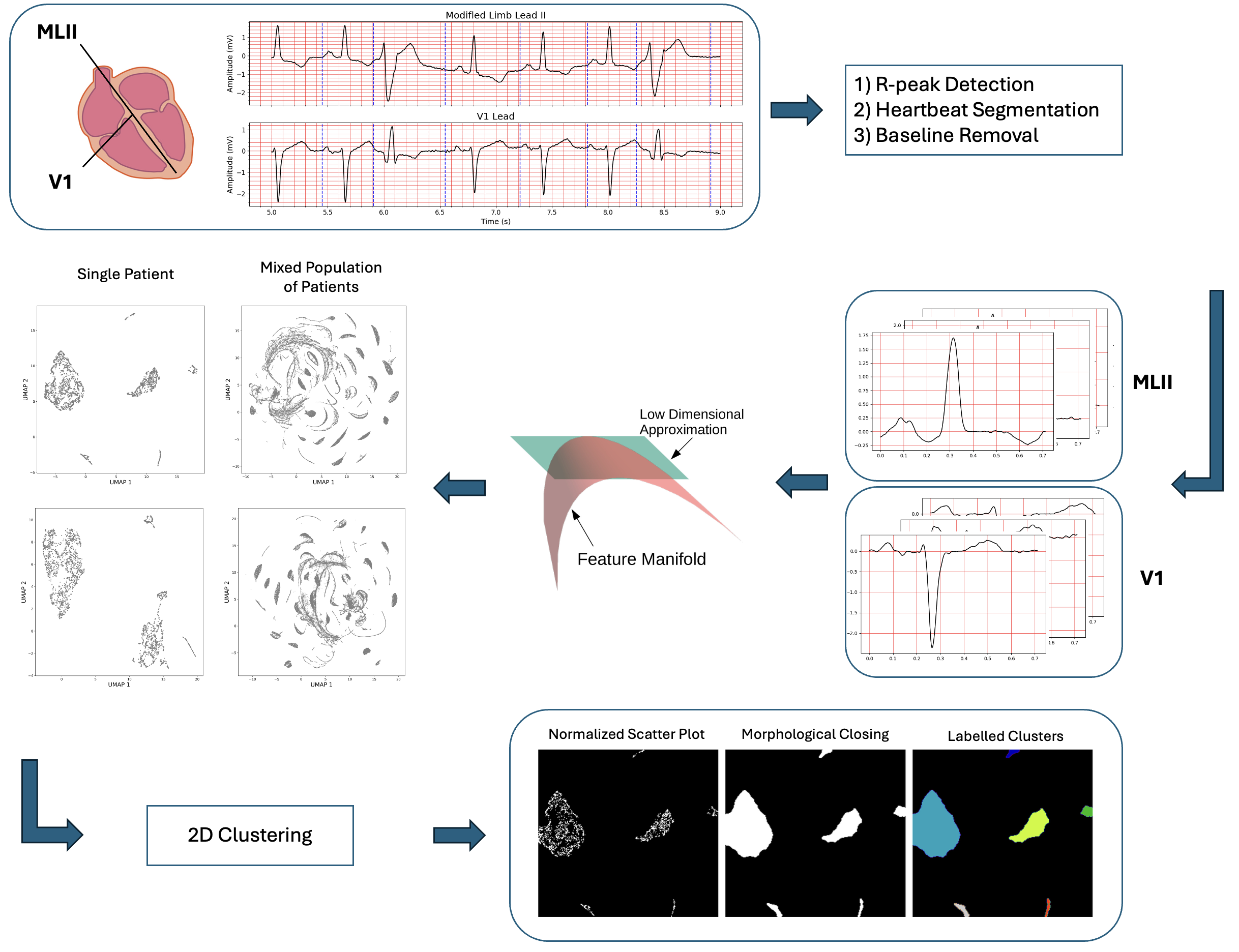} 
    \caption{Schematic of our approach.}
    \label{fig:method}
\end{figure*}

The unsupervised machine learning pipeline used in this study is illustrated in Figure~\ref{fig:method}. We first describe the dataset and labeling standard for heartbeat classification, followed by the preprocessing steps to segment individual heartbeats from the raw ECG signals and denoising. Next, we present the dimensionality reduction techniques and clustering algorithm employed, and finally the evaluation metrics used to quantify performance.

\subsection{Dataset and Labeling Overview}
We utilized the MIT-BIH Arrhythmia Database, which consists of 48 ECG recordings, each containing two channels sampled at 360 Hz and lasting slightly over 30 minutes. The dataset includes 23 recordings named 100-124 and 25 recordings named 200-234, with some numbers missing in each range. In most recordings, the first channel corresponds to modified limb lead II (MLII). Modified leads differ from standard leads in that their electrodes are placed on the chest rather than on the limbs. The second channel is a precordial lead that varies between recordings: it is most commonly V1, but in some patients it is V2, V4, or V5, depending on the subject  \cite{goldberger_physiobank_2000}. To ensure consistency across recordings, we selected only those in which MLII was used as the first lead and V1 as the second, reducing the dataset from 48 to 40 recordings. The description of the dataset can be found in Table \ref{tab:dataset_overview}.

\begin{table*}[t]
\centering
\caption{Distribution of heartbeat types according to the AAMI standard for the reduced MIT-BIH dataset.}
\label{tab:AAMI_overview}
\vspace{0.3cm}
\scriptsize
\begin{tabular}{llllr}
\multicolumn{1}{c}{\bf AAMI Superclass} & 
\multicolumn{1}{c}{\bf MIT-BIH} & 
\multicolumn{1}{c}{\bf Meaning} & 
\multicolumn{1}{c}{\bf Occurrence} & 
\multicolumn{1}{c}{\bf Percentage} \\
\hline \\
\multirow{5}{*}{\begin{tabular}[c]{@{}l@{}}Normal beats\\(79,607 beats)\end{tabular}} 
& N or . & Normal beat & 65,569 & 82.36\% \\
& L & Left bundle branch block beat & 8,072 & 10.14\% \\
& R & Right bundle branch block beat & 5,726 & 7.19\% \\
& e & Atrial escape beat & 16 & 0.02\% \\
& j & Nodal (junctional) escape beat & 224 & 0.28\% \\
\hline \\
\multirow{2}{*}{\begin{tabular}[c]{@{}l@{}}Ventricular ectopic beats\\ (7,136 beats)\end{tabular}}
& V & Premature ventricular contraction & 7,030 & 98.51\% \\
& E & Ventricular escape beat & 106 & 1.49\% \\
\hline \\
\multirow{4}{*}{\begin{tabular}[c]{@{}l@{}}Supraventricular ectopic beats \\ (2,700 beats)\end{tabular}}
& A & Atrial premature beat & 2,496 & 92.44\% \\
& S & Supraventricular premature beat & 2 & 0.07\% \\
& J & Nodal (junctional) premature beat & 52 & 1.93\% \\
& a & Aberrated atrial premature beat & 150 & 5.55\% \\
\hline \\
\multirow{3}{*}{\begin{tabular}[c]{@{}l@{}}Unknown beats\\(3,893 beats)\end{tabular}}
& Q & Unclassifiable beat & 15 & 0.39\% \\
& f & Fusion of paced and normal beat & 260 & 6.68\% \\
& / & Paced beat  & 3,618 & 92.94\% \\
\hline \\
\multirow{1}{*}{\begin{tabular}[c]{@{}l@{}}Fusion beats \end{tabular}}
& F & Fusion beat & 794 & 100\% \\
\hline \\
\multirow{8}{*}{\begin{tabular}[c]{@{}l@{}}Other beats\\(2,987 beats)\end{tabular}}
& " & Comment annotation & 437 & 14.63\% \\
& \textasciitilde & Change in signal quality & 560 & 18.74\% \\
& ! & Ventricular flutter wave & 472 & 15.80\% \\
& $|$ & Isolated QRS-like artifact & 131 & 4.39\% \\
& [ & Start of ventricular flutter/fibrillation & 6 & 0.20\% \\
& ] & End of ventricular flutter/fibrillation & 6 & 0.20\% \\
& + & Rhythm change & 1,182 & 39.57\% \\
& x & Non-conducted P-wave (blocked APB) & 193 & 6.46\% \\
\hline \\
\textbf{Total Number of Heartbeats} & & & \textbf{97,117} \\
\end{tabular}
\end{table*}

\begin{figure*}[htbp]
    \centering
    \includegraphics[width=\textwidth]{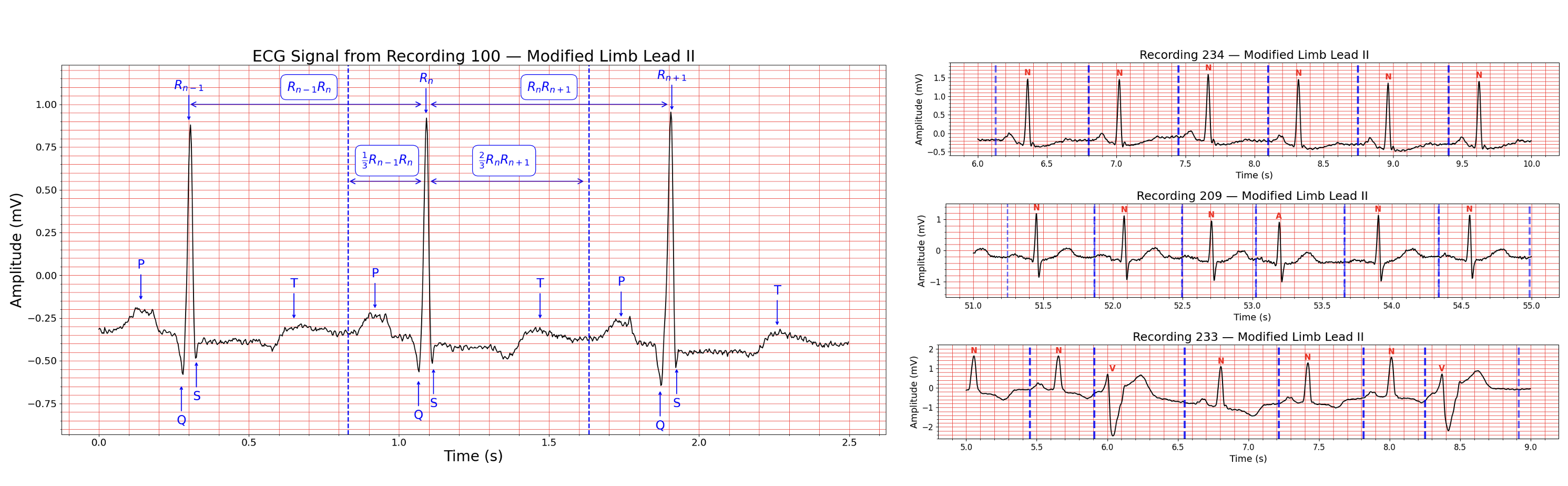} 
    \caption{Fiducial points and method of heartbeat segmentation. Each heartbeat is identified by its corresponding R-peak. The signal for each heartbeat includes the last one-third of the preceding RR-interval and the first two-thirds of the subsequent RR-interval. Three examples from different ECG strips are shown on the right, including normal beats (N), premature atrial contractions (A), and premature ventricular contractions (V), demonstrating that our segmentation performs well across different arrhythmic conditions.}
    \label{fig:Segmentation}
\end{figure*}

An ECG signal typically consists of five key fiducial points, known as P, Q, R, S, and T (Figure~\ref{fig:Segmentation}). The point P is taken as the start of the heartbeat cycle, while the R-peak is associated with the pumping of the left ventricle and is commonly used as a reference point to identify individual heartbeats. ECG datasets are typically annotated in one of two ways: beat-level, where individual heartbeats—identified by their R-peaks—are labeled (e.g., as normal or arrhythmic), or rhythm-level, where entire sequences (such as 10-second segments) are assigned a single label \cite{merdjanovska_framework_2023}. The MIT-BIH dataset follows the more detailed (and user-intensive) beat-level approach, providing over 20 distinct labels for individual heartbeats (with the overwhelming majority labeled normal); the labels with their meaning are included in Table \ref{tab:AAMI_overview}. To simplify the initial analysis, we adopted the AAMI standard for our analysis and consolidated the heartbeat types into five superclasses: normal (N), supraventricular ectopic (S), ventricular ectopic (V), fusion (F), and unknown (Q). Any heartbeat class not included in the AAMI scheme was grouped into a separate category labeled  “others” (O). This mapping, summarized in Table~\ref{tab:AAMI_overview}, consolidates rare classes (e.g., the two ventricular ectopic types) into broader categories and reduces the number of labels, resulting in a more balanced framework for analysis.

\subsection{Signal Segmentation and Filtering}
\label{Signal Segmentation and Filtering}
The first step in real-world analysis is to segment the ECG signals into isolated heartbeats. This requires identifying the R-peaks, which must be detected from raw signals. Various algorithms have been developed for this purpose, including methods by Christov \cite{christov_real_2004}, Pan-Tompkins \cite{pan_real-time_1985}, and the NeuroKit2 framework \cite{makowski_neurokit2_2021}. A recent study showed that NeuroKit2 typically performs best \cite{baraeinejad_design_2022}. Since the MIT-BIH dataset already annotates R-peak locations, we use those annotations here.  After detecting R-peaks, the signal must be segmented to establish a one-to-one correspondence between each beat and its waveform. The most straightforward approach would be to use a fixed time window, such as one second (i.e., 0.5 s before and after each R-peak), but both normal and arrhythmic heart rate variability can cause overlap between consecutive beats. This mixing contaminates the target signal and complicates downstream analysis. To overcome this, we used a constant division ratio between each R-peak: for each heartbeat, we extracted the first two-thirds of the upcoming RR interval and the last one-third of the previous RR interval (Figure \ref{fig:Segmentation}). Since this segmentation produced variable-length signals, we resampled all heartbeat waveforms to a uniform length of 256 points \textcolor{black}{(the nearest power of two to the average beat length of 267 samples) and then applied an anti-aliasing filter to avoid distortion during down-sampling.} Finally, to remove low-frequency fluctuations caused by factors such as respiration or electrode movement  \cite{baraeinejad_design_2022}, we subtracted a median-filtered version of each heartbeat waveform from the original signal. \textcolor{black}{We found that a kernel size of 127 was sufficient to remove noise from baseline wandering.}

\subsection{Dimensionality Reduction}

Heartbeats were then pairwise correlated using dimensionality reduction algorithms, which compare signal profiles in high-dimensional space and embed them in lower dimensions while preserving as much of the original graph structure as possible. The analysis was performed with three unsupervised machine learning algorithms: PCA, t-SNE, and UMAP. They were unsupervised in the sense that none of the algorithms used any labels to organize the data; rather, labels were added only after the final plotting, to try to interpret the results.
The three methods are also known as dimensionality reduction algorithms, because they start with the high dimension of the original data space (256 dimensions for each waveform) and plot the reorganized data in a lower-dimensional visualization space (2 dimensions here). PCA is the most straightforward, as it first diagonalizes the covariance matrix and then takes a subset of the largest eigenvalues; the corresponding eigenvectors give the directions of the largest variances and form the basis (axes) of the reduced latent space used for visualization. Heartbeats that are most correlated along these directions are placed closer together. However, PCA places two very severe restrictions on data representation: 1) it uses only linear algebra, and 2) it discards information from all but the largest variations. Both t-SNE and UMAP begin with the covariance matrix as well (and are often initialized with the results of PCA, as done here), but they use nonlinear mappings to organize the data. t-SNE assumes a Gaussian distribution of points in the original data space but a Cauchy distribution in the visualization space, with “attractive” forces grouping data with similar features into clusters and “repulsive” forces separating clusters \cite{JMLR:v9:vandermaaten08a}. UMAP extends this technique by considering local density variations among the original data points, i.e., by defining a Riemannian metric in the high-$D$ space, and modifying the low-$D$ space with two extra parameters to adjust the width and fall-off of the t-distribution \cite{mcinnes_umap_2018}. In both cases, the mapping to the lower-dimensional manifold is obtained by minimizing the cross-entropy between it and the target distribution. 

\textcolor{black}{
Both t-SNE and UMAP include hyperparameters that can influence the resulting embeddings and the clustering performed on the 2D projections. For all results presented in this study, UMAP was configured with \texttt{n\_neighbors=15} and \texttt{min\_dist=0.1}, and t-SNE with a perplexity of \texttt{30}, both using PCA initialization and fixed random seeds for reproducibility. All remaining parameters were kept at their default values. The parameter \texttt{n\_neighbors} governs the trade-off between local and global structure in the embedding, while \texttt{min\_dist} controls the compactness of points in the low-dimensional space.}

\subsection{2D Clustering Algorithm}
After projecting the signals into a 2D latent space using UMAP or t-SNE, clustering is required to separate distinct groups. Manual clustering is slow and non-automated, while methods such as k-means require specifying the number of clusters in advance. Density-based algorithms such as DBSCAN \cite{DBSCAN} and HDBSCAN \cite{hutchison_density-based_2013} address this limitation, but remain sensitive to other hyperparameter selection. In practice, UMAP and t-SNE form visually separable clusters in two dimensions. Here, we treat the scatter plot as an image and apply a simple image-processing pipeline: after adjusting the resolution and aspect ratio, morphological area closing fills small gaps between points, and connected components are extracted to identify clusters (\textcolor{black}{if a cluster contains fewer than a predefined number of samples, its members are reassigned to the nearest cluster.}) The resulting algorithm is robust to irregular and elongated clusters commonly observed in NLDR embeddings. A comparison of this approach with standard clustering methods is shown in the Supplementary Material.

\subsection{Evaluation Metrics}

Many metrics can be used to evaluate the performance of dimensionality reduction and clustering algorithms. To quantify the former, we use the trustworthiness score, which measures how well local neighborhood relationships from the high-dimensional space are preserved in the low-dimensional embedding. Trustworthiness ranges from 0 to 1, where 1 indicates perfect preservation of local neighborhoods \cite{goos_neighborhood_2001}. \textcolor{black}{This metric evaluates the performance of dimensionality reduction independently, before any clustering is applied in the 2D space.} 

\textcolor{black}{To evaluate the quality of clustering, we report the Silhouette coefficient, Davies–Bouldin index, and Calinski–Harabasz index. These metrics measure cluster cohesion and separation based on the intrinsic structure of the data, independent of external labels, and are computed after applying the clustering algorithm to the 2D embedding space \cite{arbelaitz2013}. The first score ranges from $[-1,1]$ (with 1 the best value), while the latter indices can take on any value greater than zero (with values closer to zero indicating higher quality for D-B and lower quality for C-H). If the clusters are convex and non-nested, then the first two metrics are more informative and reliable \cite{chicco2025}. } 

None of these metrics, however, assess the usefulness of the embeddings for downstream tasks. A more task-oriented approach is the application of a simple K-Nearest Neighbors classifier (with varying values of $K$) to compare the predicted labels with the ground truth \cite{mcinnes_umap_2018}. This approach evaluates whether samples with the same labels are close to each other and separated from those with different labels, while also quantifying how informative the features are for downstream classification. \textcolor{black}{Historically, k-NN has been applied only in the low-dimensional space, as decisions about categories and subsequent analysis are typically based on the reduced mapping. Here, we perform the same k-NN analysis in the original high-dimensional space ($\mathbf{R}^{256}$) as well, to compare the classification performance of the embeddings with that of the original representation.}

\section{Results}
We first discuss the embeddings obtained when dimensionality reduction algorithms are applied to the entire population, i.e. to heartbeats from all subjects. We then analyze embeddings of individual subjects. \textcolor{black}{To demonstrate the versatility of the methods with respect to electrode configuration and cardiac condition, we focus here on recordings 102, 116, 207, 209, and  231. Additional recordings are analyzed in the Supplementary Material.}

\subsection{Dimensionality Reduction of Mixed Populations}
Figure~\ref{fig:MIT-BIH_nldr_all} shows the results of PCA, t-SNE, and UMAP applied to the mixed population of heartbeats. \textcolor{black}{We first plot the raw mappings, which sort the waveforms based solely on their similarities and differences. To evaluate algorithm performance and aid interpretation, we then label the points by the MIT-BIH metadata, i.e. by AAMI class, patient number, and gender.}

\subsubsection{PCA}
We begin with the PCA projection of lead II, which aligns with the heart’s long axis and captures electrical activity from atrial initiation to ventricular activation and recovery. The unlabeled mapping is shown in the first column. Most of the data is placed in a central, crab-like structure, with the rest placed in diffuse point clouds around the periphery. These latter points have more variation than those in the main structure, suggesting that their waveforms are more anomalous. This intuition is partially confirmed in the second column, which shows the same points labeled by AAMI category. While the lower clouds consist of distinct classes (V and U), the upper clouds do not (there is a partial diagonal cloud of ventricular beats at the left of the main structure, but half of its points are hidden in the unlabeled image). The S and O categories are hidden within the main cluster, as are the remaining ventricular beats. The “other” category, colored in cyan, is very close to the origin, indicating almost no variation along the first two principal components; indeed, examining these signals confirms that their waveforms are nearly flat, likely due to electrodes detaching during measurement. Attempts to isolate anomalies with less variation in the PCA plot fail convincingly. For example, heartbeats in the prominent nodes at the corners of the “crab” and in the bridge between its upper “claws” are all normal.  The ambiguity is even worse when considering patient identifiers. For the recording label, only the lower right clouds correspond to heartbeats from the same individual. These patients are both male. Interestingly, the extreme heartbeats on both the left and right are also primarily from male patients, while the gender contributions are roughly equal in the outlying bottom and top clusters.

Some of these heartbeats become more distinct in the V1 lead, which measures the same electrical signals from an orthogonal point of view. As before, we first consider only the unlabeled data shown in the first column. The most distinct clusters are in the top of the figure, with a T-shaped cloud of points in the upper left and a pair of similarly shaped clusters in the upper right. There are also several less distinct groups that are fragmented from the central “pangea” of points, including one on top, one to the left, and one to the lower left. Labeling with AAMI metadata reveals that the T-shaped cluster consists of two beat types: F beats in the stem and V beats in the crossbar. In the upper right, the larger cluster of the pair consists of normal beats, while the smaller contains V beats. As the dominant ventricular arrhythmias are premature ventricular contractions (PVCs), the relative placements of these are consistent with near-atrial synchronization with fusion beats and heartbeats that appear more normal but fire early. Interestingly, the T-shaped heartbeats appear to belong to the same, male patient, while the pair in the upper right come from female patients. Moving counterclockwise on the main structure, the top cluster consists of normal heartbeats primarily from the same female patient, the left cluster has normal beats from two male patients, and the beats in the lower left cluster come mostly from a single male patient.

\subsubsection{t-SNE and UMAP}
The nonlinear mappings of t-SNE and UMAP are dramatically different than the linear one of PCA. Most significantly, much of the data is concentrated into distinct clusters, with UMAP forming tighter clusters that are more separated from each other. The remaining points are localized to large arcs, with UMAP placing them in roughly continuous islands and t-SNE placing them in more discretized archipelagos. Overall, the distribution of t-SNE is circular, a result of its initial attempt to model the data with a single Gaussian distribution. The UMAP visualization has more of a grid pattern, inherited from its choice of multiple Gaussians and its attempt to approximate a uniform manifold; the orientation of the grid stems from its initialization with PCA.

Labeling the points shows the superior action of nonlinear dimension reduction. After assigning AAMI labels, most arrhythmic beats tend to cluster near each other; however, distinguishing between them without labels remains challenging. Notably, two distinct clusters of unknown beats (yellow) are consistently observed across both leads, and a prominent red cluster, corresponding to supraventricular ectopic beats, also emerges in all visualizations. When visualized by patient ID, a striking pattern appears: many of the distinct clusters correspond to different individuals. This suggests that both UMAP and t-SNE find biometric signals from the same person to be a stronger attractor than any characteristics of the heartbeat categories. That is, these algorithms effectively perform a patient identification task rather than an arrhythmia classification task (an observation that holds even when considering cardiac differences in gender \cite{st_pierre_sex_2022, prajapati_sex_2022}, as shown in column 4). While prior studies have demonstrated ECG-based identification using supervised learning \cite{biel_ecg_2001, LI_ETUMAP_2022}, our results reveal individuality in an unsupervised manner. 

\subsubsection{Quantitative Results}
\begin{figure*}[htbp]
    \centering
    \includegraphics[width=0.83\textwidth]{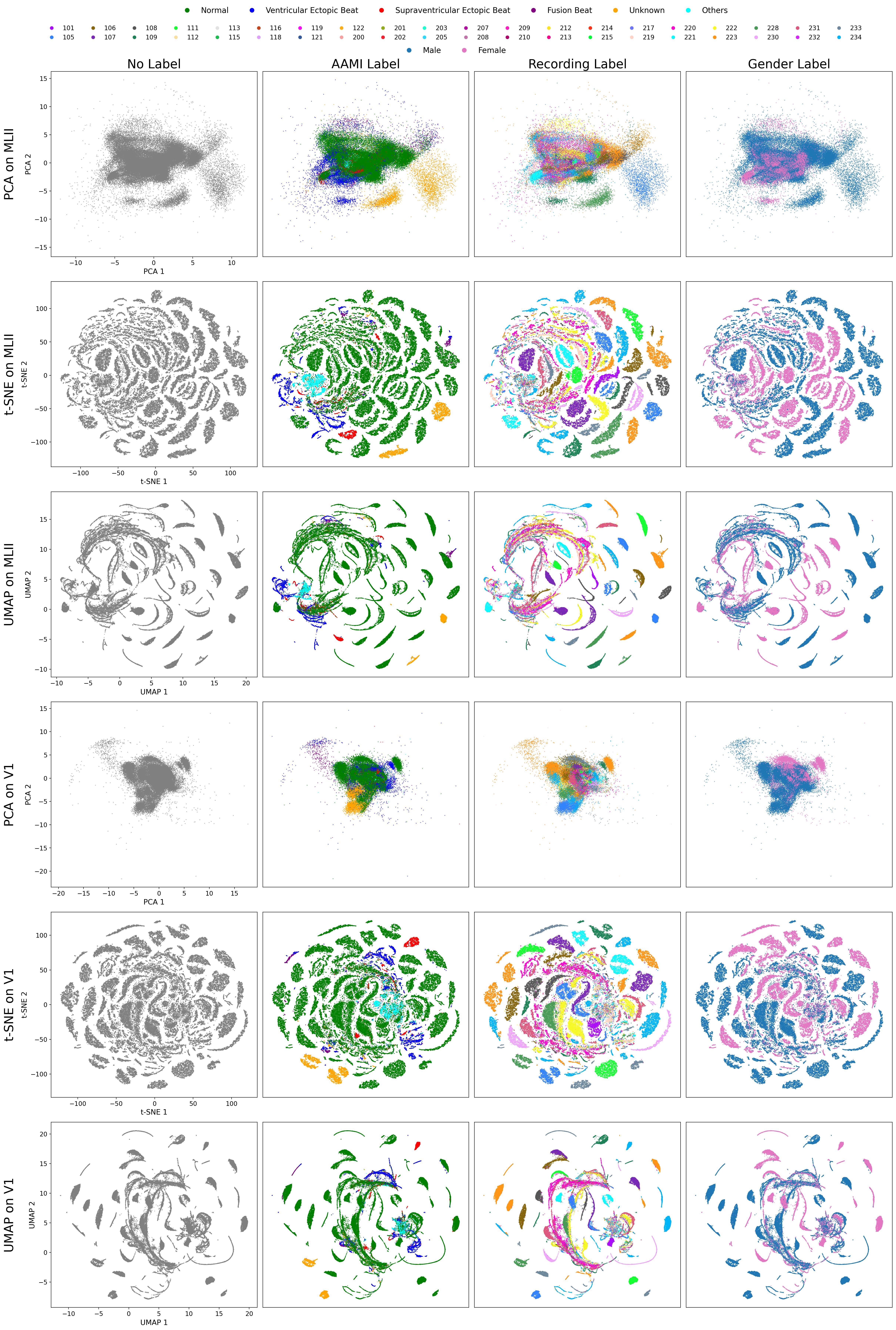} 
    \caption{Visualization of heartbeat signal populations from 40 recordings. Shown are the 2D latent spaces using PCA, t-SNE, and UMAP from the MLII lead (top set) and the V1 lead (bottom set). Column 1 shows projections without any labels. The subsequent columns show data labeled with (2) heart arrhythmia types according to the AAMI standard, (3) patient recording number, and (4) gender.}
    \label{fig:MIT-BIH_nldr_all}
\end{figure*}

\begin{figure}[t]
\centering
\includegraphics[width=\columnwidth]{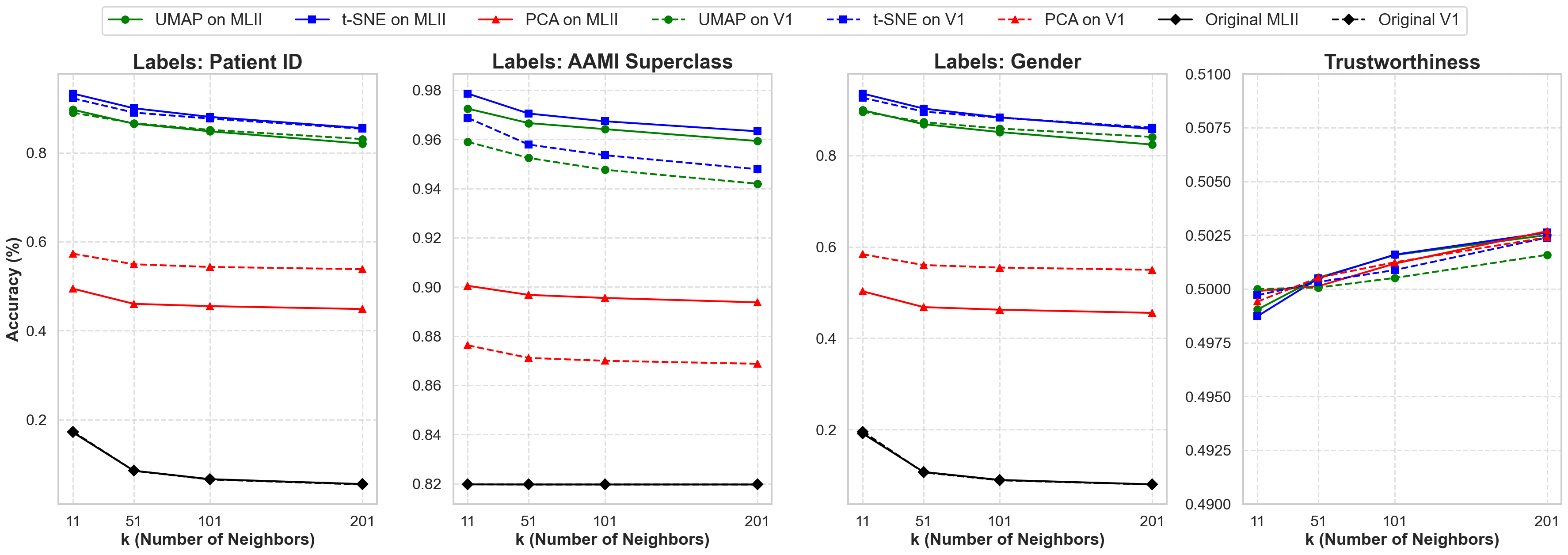}
\caption{\textcolor{black}{k-NN classification and trustworthiness results based on the embeddings shown in Figure~\ref{fig:MIT-BIH_nldr_all}.}}
\label{fig:identification_metric}
\end{figure}

To quantitatively assess the usefulness of the embeddings for downstream classification tasks, e.g., patient identification, arrhythmia detection, and gender classification, we applied a k-NN classifier (with $k \in \{11, 51, 101, 201\}$) to the different 2D embeddings in Figure~\ref{fig:MIT-BIH_nldr_all}, and reported classification accuracy. The results, summarized in Figure~\ref{fig:identification_metric}, show that the nonlinear mappings significantly outperformed PCA in all cases. Both nonlinear methods achieved comparable accuracy, with t-SNE showing slightly higher numerical scores. However, UMAP produced wider separation between neighboring clusters, which can facilitate label-free clustering. The accuracy for arrhythmia detection exceeded 95\% for both t-SNE and UMAP; however, this requires after-the-fact labeling and cannot be determined from visual inspection alone. Most of the distinct clusters corresponded to different patients, and the results indicate that patient identification can be achieved with an accuracy of $\geq 80\%$. \textcolor{black}{Notably, applying k-NN directly to the original high-dimensional signal space yields substantially lower accuracy in all cases, indicating that the embeddings effectively reduce dimensionality while preserving structure by grouping similar signals together and separating dissimilar ones.} 

We also calculated trustworthiness to assess how well local neighborhoods are preserved in the lower-dimensional space and found that all algorithms yielded similarly low scores around 0.5 when applied to the mixed population. This is probably because multiple sources of variation, such as individual differences, arrhythmia types, and gender, cannot be fully captured in two dimensions without distortion.

\subsection{Personalized Results}
We next applied PCA, t-SNE, and UMAP to heartbeat profiles from individual recordings. Results from \textcolor{black}{five} patients are presented in Figures \ref{fig:signal_examples116} -- \ref{fig:signal_examples102} (The results for all recordings are provided in Supplementary Material). As before, both UMAP and t-SNE effectively separated signal phenotypes into distinct clusters. While most of these are obvious to identify by hand, the ultimate goal is to demarcate them automatically in an unsupervised manner. To this end, we applied the clustering algorithm described in the Methodology section to make cluster domains. Each patient can have different numbers of clusters, which vary depending on the number and type of normal vs. abnormal heartbeats; for ease of discussion, we labeled the clusters for each recording in order of largest to smallest. Representative heartbeats for each class, along with their averages and variance, are given in Figures \ref{fig:signal_examples116} -- \ref{fig:signal_examples102}. Finally, we labeled each cluster according to its AAMI super-class and MIT-BIH sub-class. 

\subsubsection{Recording 116}
Results from Recording 116 are shown in Figure \ref{fig:signal_examples116}. Applying PCA on both leads MLII and V1 gives radial clusters that are distinctly separated, with a small cloud of points centered at their common origin. UMAP on MLII again results in the best cluster separation, and to interpret the results further, we concentrate on these visualizations only. We begin by looking at AAMI super-class labels. Interestingly, there are three distinct clusters of heartbeats labeled as normal: the largest one (Cluster 1), exhibiting an elliptical shape, and the next two, which are almost linear clusters that are perpendicular to each other (Clusters 2 and 3). Reasons for this separation are evident in their representative waveforms. We will take cluster 1 as the “most” normal, i.e. the standard reference: it has a small P wave, a large, thin R-peak with small S dip, a prominent, rounded T wave, and a long end recovery at zero voltage. Cluster 2 shares the same features and timing, but its R and T peaks have significantly lower amplitudes, with an S dip that is much more prominent. Cluster 3 has a broad, moderate-amplitude R peak that occurs later than the other normals and a T falloff that continues deeply into negative voltage. Cluster 4 in the lower left starts off deeply negative and has a combined P-R complex (with nearly twice the amplitude of cluster 3), a negative T wave, and an extra-long refractory period at zero voltage. Both the R and T waves occur earlier than the other normal, consistent with the dominance of PVCs in this cluster. Finally, the smallest cluster 5 has heartbeats that average to a near-constant zero. This anomalous behavior is noted by both the AAMI and MIT-BIH systems and can originate from flatline or random waveforms, indicating changes in signal quality such as electrode disconnection.

\subsubsection{Recording 231}
We next consider Recording 231 (Figure \ref{fig:signal_examples231}). This patient has four major clusters of heartbeats (more evident in the UMAP embedding of the V1 lead), again with three of them labelled normal by the AAMI classification (1, 3, and 4), \textcolor{black}{along with a small, compact cluster (5) of normal beats.} As before, the “other” cluster (2 here) is nearly flatline. This time, the MIT-BIH labeling distinguishes two classes of normal: Clusters 1 and 4 are characterized as right bundle branch block (RBBB) beats, while the remaining cluster (3) is classically normal. The RBBB beats are most easily classified by their V1 profiles, which show a tell-tale “bunny ear” profile caused by a distinct r-like sub-peak before the classical QRS complex \cite{ikeda_right_2021}. In this case, the two clusters are distinguished by the different timings of these complexes and the dramatic difference in the refractory period at the end of the waveform. \textcolor{black}{Clusters 3 and 5 are very similar, with the latter displaying more prominent peaks in the ventricular part of the beat.}

\subsubsection{Recording 209}
Recording 209, shown in Figure \ref{fig:signal_examples209}, is more complex, with 5 separate clusters on the MLII lead. The largest cluster consists of normal beats, while Clusters 2, 3, and 5 are dominated by premature atrial beats. Despite the common label, the latter clusters have two very distinct behaviors. \textcolor{black}{For example, Cluster 3 exhibits a much more pronounced T-peak compared to Clusters 2 and 5.} Overall, the waveforms in Cluster 3 are similar to those in Cluster 1 (thus some joint placement) but with subdued R- and T-peaks that fire earlier than normal beats; there are also differences in polarization behavior at the beginning and end of each cycle. In contrast, the waveforms in Clusters 2 and 5 have a dispersed P profile, stronger R-peak amplitudes, weaker and longer T pulses, and a prominent U wave; all these are signs of severe hypercalcemia \cite{alzate_electrocardiographic_2015}. Finally, Cluster 4 contains a grab-bag of “other” beats, such as changes in rhythm and signal quality, and isolated QRS artifacts, which average to nearly zero variation.

\subsubsection{Recording 207}
A similar variety of "other" beats occurs in the 2nd largest cluster of Recording 207, shown in Figure \ref{fig:signal_examples207}. In this case, the dominant beat is ventricular flutter, with elements of left and right bundle branch blockages, ventricular escape, and premature ventricular contraction. 
\textcolor{black}{Clusters 1 and 3 appear very similar at first glance, but the latter shows delayed R and T peaks and a broader R profile. These subtle differences are captured by NLDR and carry clinical significance, as reflected in their differing AAMI labels; Cluster 1 is considered normal in AAMI, while Cluster 3 represents premature atrial beats. Cluster 5 is also considered normal in AAMI, but the NLDR algorithm clearly identifies its waveforms as separate from those of Cluster 1; their medical distinction is revealed in the MIT-BIH labeling, where Cluster 1 consists primarily of LBBB while Cluster 5 includes mostly RBBB.} Interestingly, this patient exhibits nearly all types of arrhythmias, including half the MIT-BIH labels and the appearance of less common Osborn/J waves (after the QRS complex in Cluster 5) \cite{shiozaki_case_2021}. Both UMAP and t-SNE are able to form distinct clusters for these different beat types in a label-free approach.

\subsubsection{Recording 102}
\textcolor{black}{Unlike most recordings in MIT-BIH dataset that contain the MLII and V1 leads, this patient’s recording includes V2 and V5 leads. This provides another opportunity to demonstrate that NLDR algorithms can reveal differences in ECG signals independent of the specific leads used. For example, applying UMAP to the V5 lead reveals three primary clusters. Cluster 1 is composed primarily of paced beats, with a small number signifying a fusion of paced and normal beats (both of which are categorized as "Unknown" under the AAMI labeling scheme). Cluster 2 primarily contains normal beats. Cluster 3 consists of a heterogeneous mixture of arrhythmias, including premature ventricular contractions, paced beats, and rhythm changes. This diversity is evident in the cluster’s average waveform, which appears nearly flat, and its relatively broad variance profile.}

\subsubsection{Quantitative Results}
\begin{figure}[t]
    \centering
    \includegraphics[width=\columnwidth]{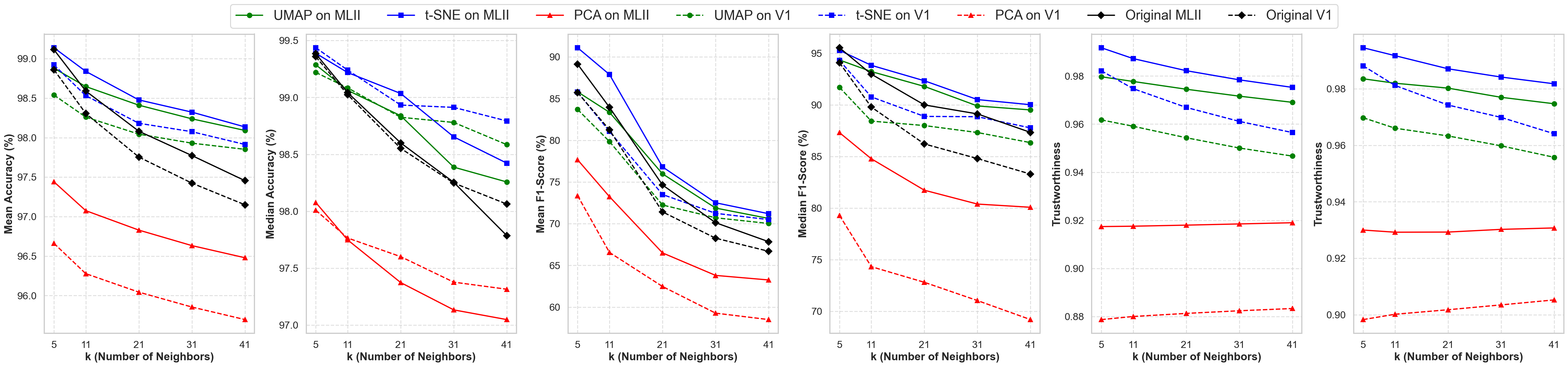}
    \caption{k-NN classification and trustworthiness results for personalized setting. A k-NN classifier with $k \in \{5, 11, 21, 31, 41\}$ was applied to both 2D embeddings and \textcolor{black}{original signals} to distinguish normal from arrhythmic beats. Accuracy, F1 scores, and trustworthiness were computed for each recording and summarized by their mean and median across 40 recordings.}
    \label{fig:classification_metric}
\end{figure}

\textcolor{black}{Figure \ref{fig:classification_metric} shows k-NN classification and trustworthiness results for all 40 individual recordings. Both t-SNE and UMAP achieved trustworthiness values $\geq 0.95$, with t-SNE’s slightly higher score indicating better preservation of local neighborhoods. Both methods substantially outperform PCA. Compared to the NLDR results from the mixed population of heartbeats (Figure~\ref{fig:identification_metric}), trustworthiness is much higher in the personalized setting. This improvement arises because focusing on a single individual eliminates variations due to different people or gender, leaving only arrhythmia-type differences as the main source of variation.} 

\textcolor{black}{In addition, clustering performance metrics, including the silhouette coefficient, Davies--Bouldin index, and Calinski--Harabasz index, are reported in Table~\ref{tab:clustering_classification_results}. These metrics consistently indicate that clustering in the UMAP embedding outperforms that in t-SNE, reflecting the tighter and more compact cluster structure that UMAP produces. To evaluate the usefulness of embeddings for downstream classification, we applied k-NN with $k \in \{5, 11, 21, 31, 41\}$ on both the 2D embeddings and the original high-dimensional space ($\mathbf{R}^{256}$), performing binary classification (normal vs.\ abnormal). Accuracy and F1 scores were computed for each individual, with the mean and median across the 40 recordings shown in Figure \ref{fig:classification_metric}. Both quantitative and qualitative analyses indicate that the MLII lead achieves higher accuracy and forms more distinct clusters compared to the V1 lead. As before, t-SNE and UMAP outperform PCA, with t-SNE showing slightly higher classification performance. However, the tighter and more compact clusters in the UMAP embedding facilitate label-free clustering.  Comparing k-NN performance in the original high-dimensional space with that in the low-dimensional embeddings shows that both t-SNE and UMAP achieve higher classification metrics than the original $\mathbf{R}^{256}$ signals, especially for larger values of $k$. This indicates that these methods do more than reduce dimensionality; they reorganize the data so that signals of the same class are more compact and better separated from other classes.}

\subsubsection{Hyperparameter Analysis}
\begin{figure*}[htbp]
    \centering

    \begin{minipage}{\textwidth}
        \centering
        \begin{overpic}[width=0.85\linewidth]{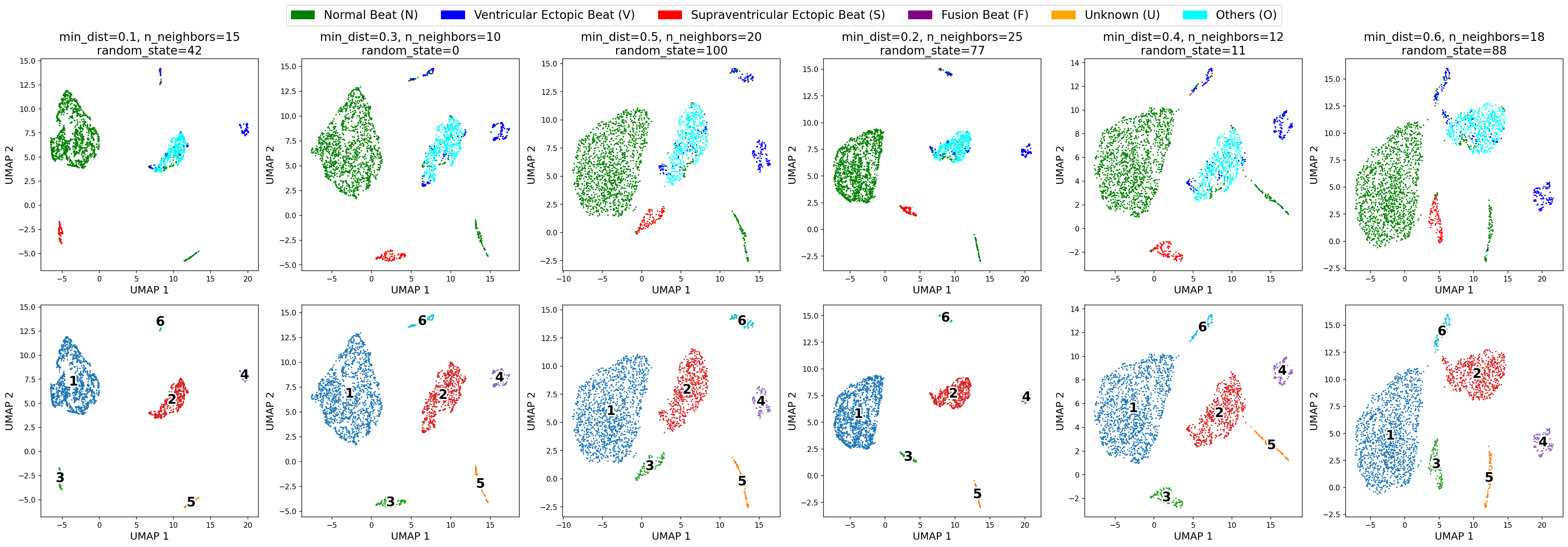}
            \put(-1.2,21){\rotatebox{90}{\tiny AAMI Labels}}
            \put(-3.5,10){\rotatebox{90}{\tiny Recording 207 - UMAP on MLII}}
            \put(-1.2,7){\rotatebox{90}{\tiny Clustering Labels}}
        \end{overpic}
    \end{minipage}

    \vspace{0.6em}

    \begin{minipage}{\textwidth}
        \centering
        \begin{overpic}[width=0.85\linewidth]{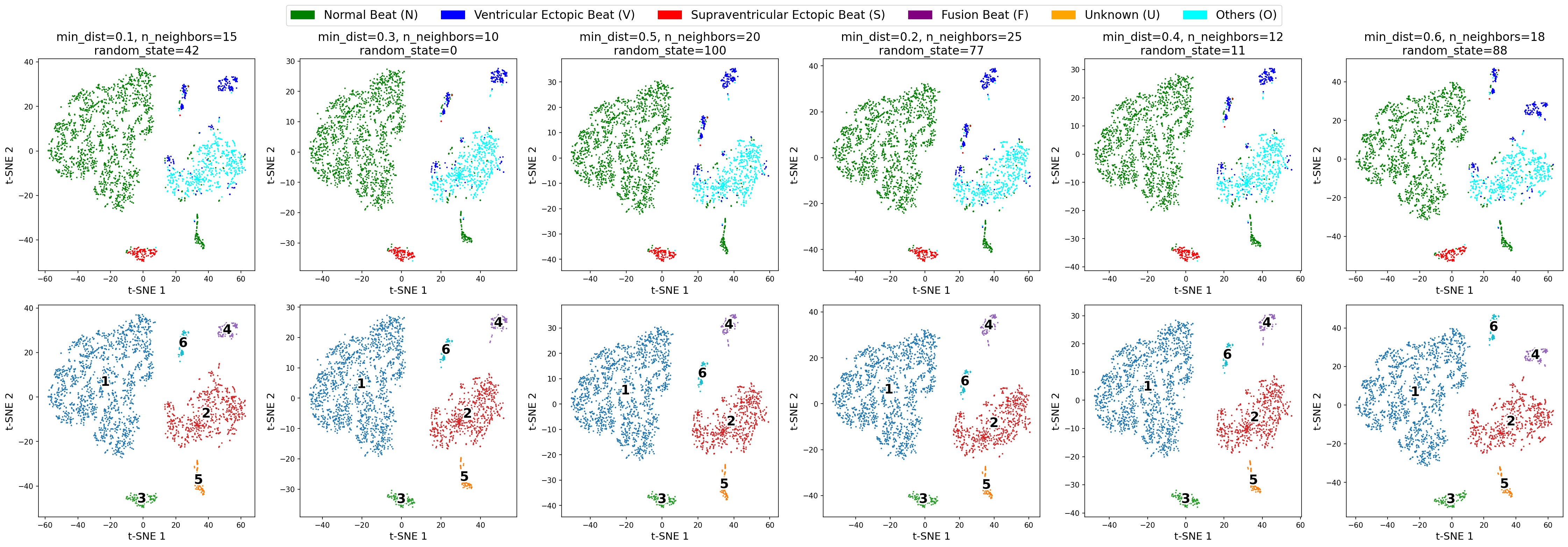}
            \put(-1.2,21){\rotatebox{90}{\tiny AAMI Labels}}
            \put(-3.5,10){\rotatebox{90}{\tiny Recording 207 - t-SNE on MLII}}
            \put(-1.2,7){\rotatebox{90}{\tiny Clustering Labels}}
        \end{overpic}
    \end{minipage}

    \vspace{0.6em}

    \begin{minipage}{\textwidth}
        \centering
        \begin{overpic}[width=0.85\linewidth]{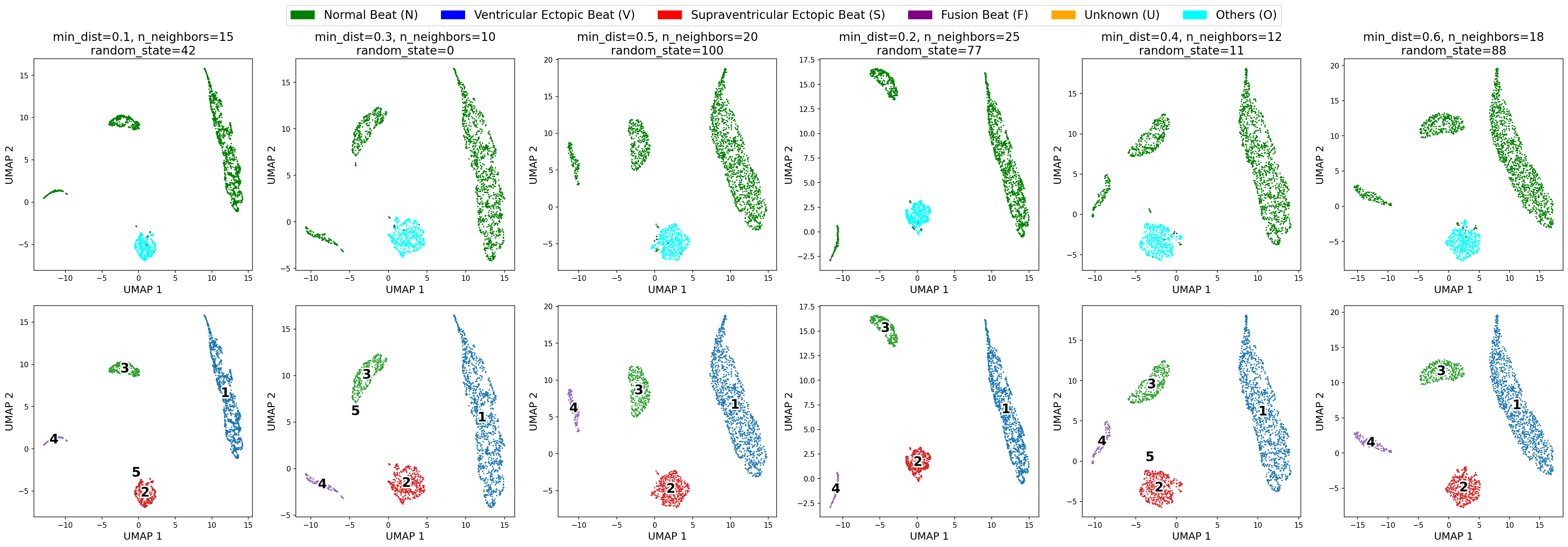}
            \put(-1.2,21){\rotatebox{90}{\tiny AAMI Labels}}
            \put(-3.5,10){\rotatebox{90}{\tiny Recording 231 - UMAP on V1}}
            \put(-1.2,7){\rotatebox{90}{\tiny Clustering Labels}}
        \end{overpic}
    \end{minipage}

    \vspace{0.6em}

    \begin{minipage}{\textwidth}
        \centering
        \begin{overpic}[width=0.85\linewidth]{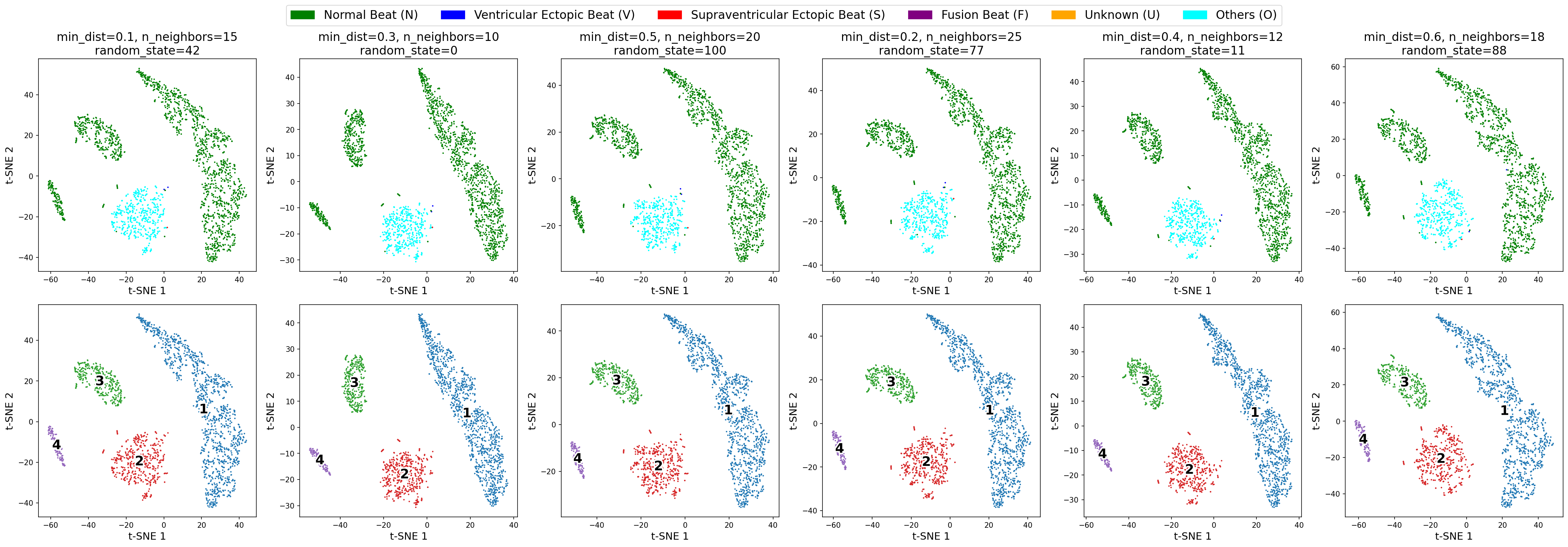}
            \put(-1.2,21){\rotatebox{90}{\tiny AAMI Labels}}
            \put(-3.5,10){\rotatebox{90}{\tiny Recording 231 - t-SNE on V1}}
            \put(-1.2,7){\rotatebox{90}{\tiny Clustering Labels}}
        \end{overpic}
    \end{minipage}

    \caption{\textcolor{black}{UMAP and t-SNE embeddings of ECG recordings 207 and 231 under different hyperparameter settings. The first row in each panel shows the AAMI labels, while the second row shows the clusters obtained using the proposed clustering algorithm. Despite changes in hyperparameters, the clusters remain visually separable.}}
    \label{fig:hyperparameters_clustering}
\end{figure*}

\textcolor{black}{To assess sensitivity to hyperparameter choices, we analyzed UMAP embeddings of the MLII signal for record 207 and the V1 signal for record 231. As shown in Figure~\ref{fig:hyperparameters_clustering}, all tested configurations yield well-separated clusters, with only minor variations in cluster orientation or inter-cluster distances. Across all configurations, the proposed clustering algorithm consistently recovers these clusters. Extended analyses of additional records, embeddings, and alternative clustering algorithms are provided in the Supplementary Material.}

\begin{figure*}[htbp]
    \centering
    \includegraphics[width=0.9\textwidth]{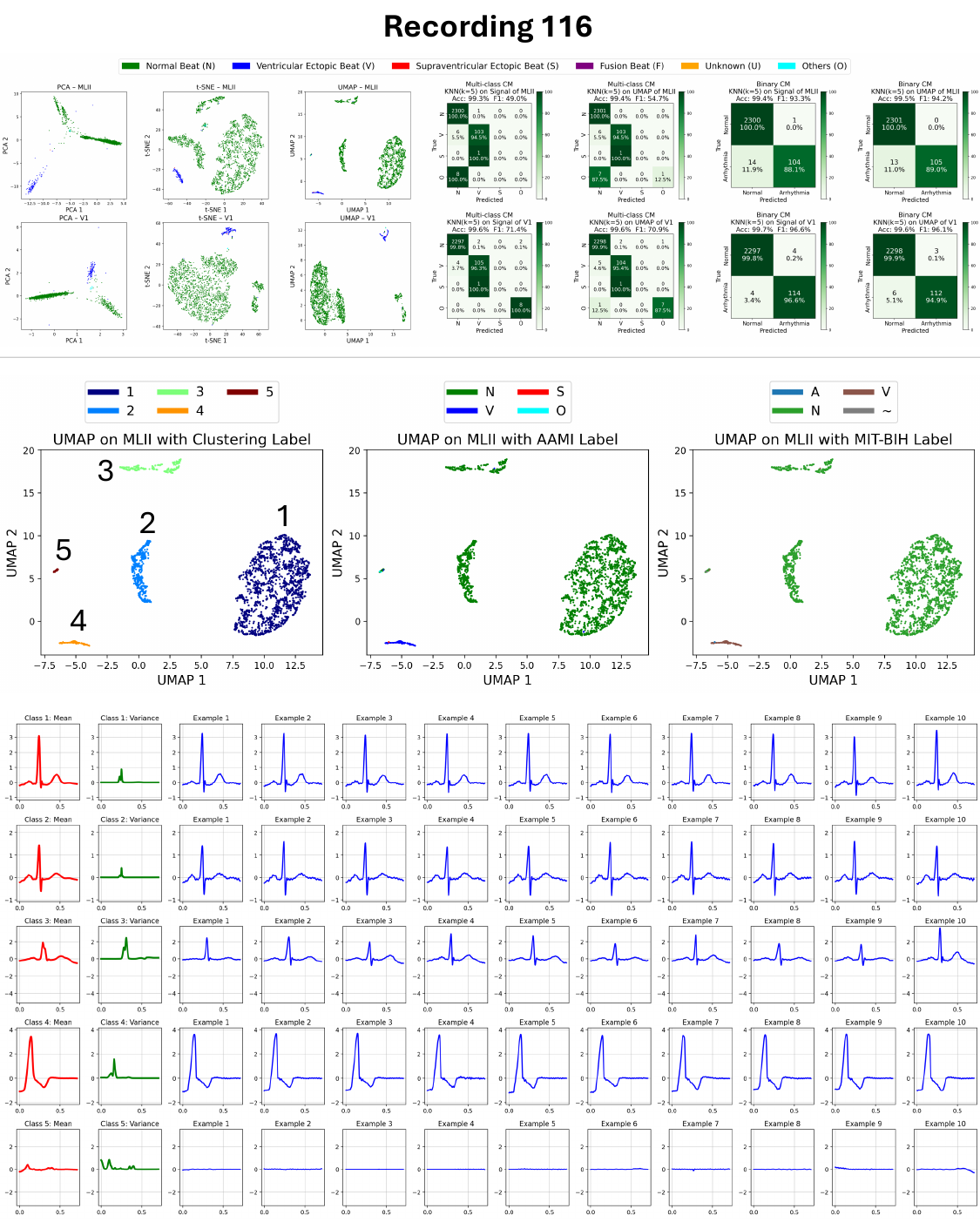}
    \caption{
    Analysis of Recording 116 with dimensionality reduction methods. 
    Top: 2D visualizations using PCA, t-SNE, and UMAP. A k-NN classifier ($k=5$) is applied on the UMAP embeddings \textcolor{black}{and on the original signals} to evaluate the 2D latent space for downstream classification. 
    Bottom: Clustering of 2D UMAP embeddings from the MLII lead, followed by labeling of heartbeats. Ten example signals per cluster are shown, along with their mean and variance. For all AAMI labels, refer to the topmost legend for character definitions. For MIT-BIH labels, see Table~\ref{tab:AAMI_overview}.}
    \label{fig:signal_examples116}
\end{figure*}

\begin{figure*}[htbp]
    \centering
    \includegraphics[width=0.9\textwidth]{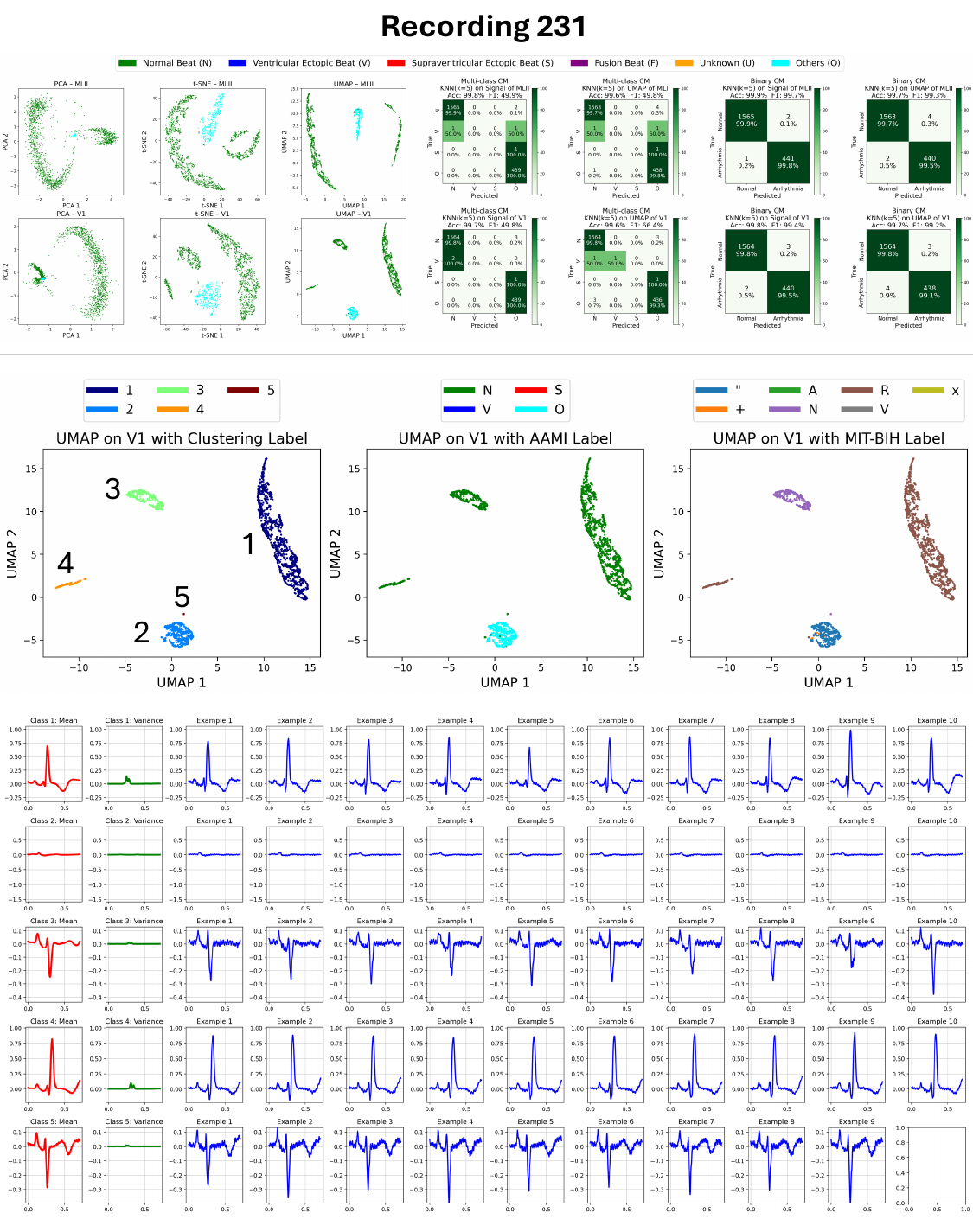}
    \caption{
    Analysis of Recording 231 with dimensionality reduction methods. 
    Top: 2D visualizations using PCA, t-SNE, and UMAP. A k-NN classifier ($k=5$) is applied on the UMAP embeddings \textcolor{black}{and on the original signals} to evaluate the 2D latent space for downstream classification.   
    Bottom: Clustering of 2D UMAP embeddings from the V1 lead, followed by labeling of heartbeats. Ten example signals per cluster are shown, along with their mean and variance. For all AAMI labels, refer to the topmost legend for character definitions. For MIT-BIH labels, see Table~\ref{tab:AAMI_overview}.}
    \label{fig:signal_examples231}
\end{figure*}

\begin{figure*}[htbp]
    \centering
    \includegraphics[width=0.9\textwidth]{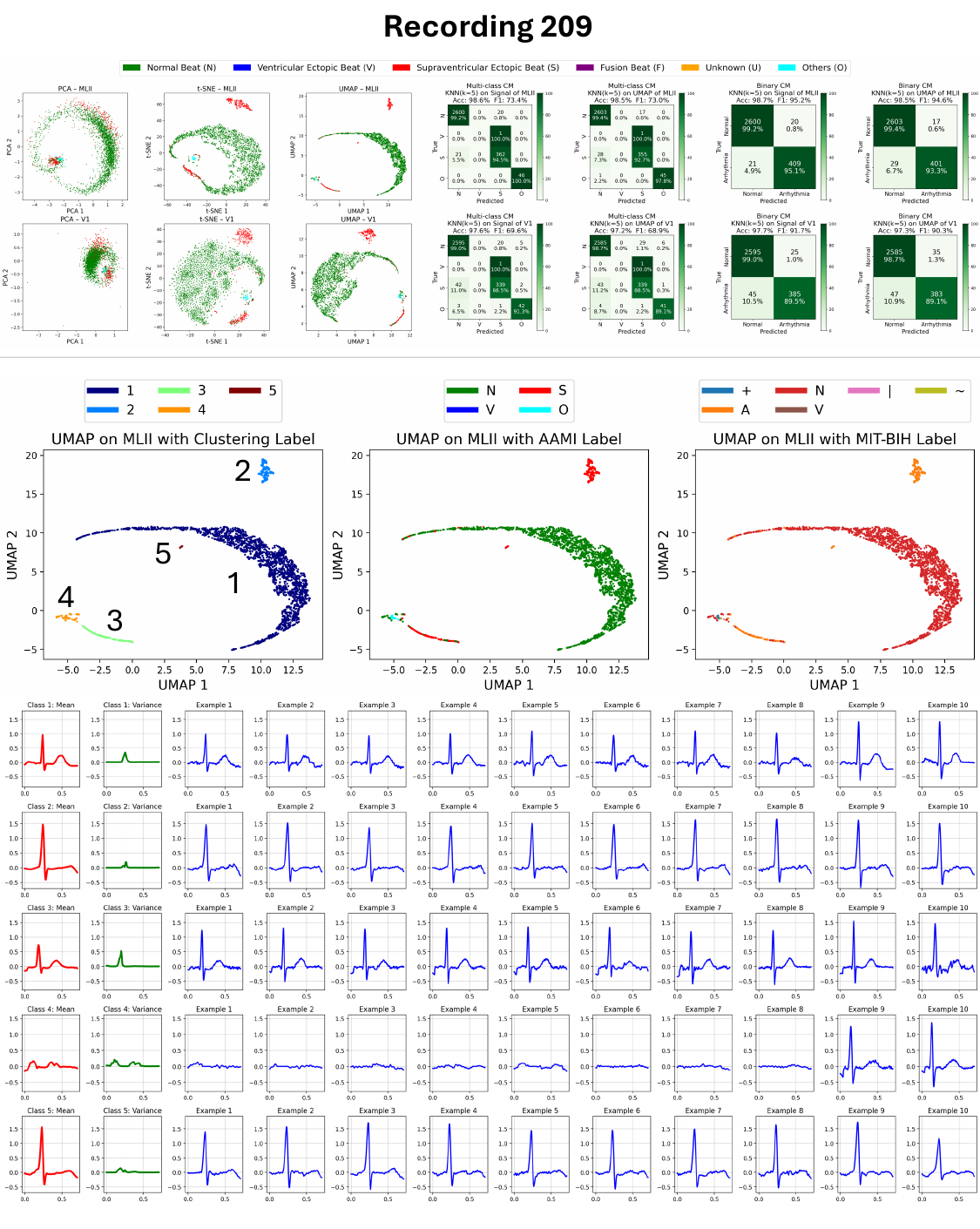}
    \caption{
    Analysis of Recording 209 with dimensionality reduction methods. 
    Top: 2D visualizations using PCA, t-SNE, and UMAP. A k-NN classifier ($k=5$) is applied on the UMAP embeddings \textcolor{black}{and on the original signals} to evaluate the 2D latent space for downstream classification.   
    Bottom: Clustering of 2D UMAP embeddings from the MLII lead, followed by labeling of heartbeats. Ten example signals per cluster are shown, along with their mean and variance. For all AAMI labels, refer to the topmost legend for character definitions. For MIT-BIH labels, see Table~\ref{tab:AAMI_overview}.}
    \label{fig:signal_examples209}
\end{figure*}

\begin{figure*}[htbp]
    \centering
    \includegraphics[width=0.89\textwidth]{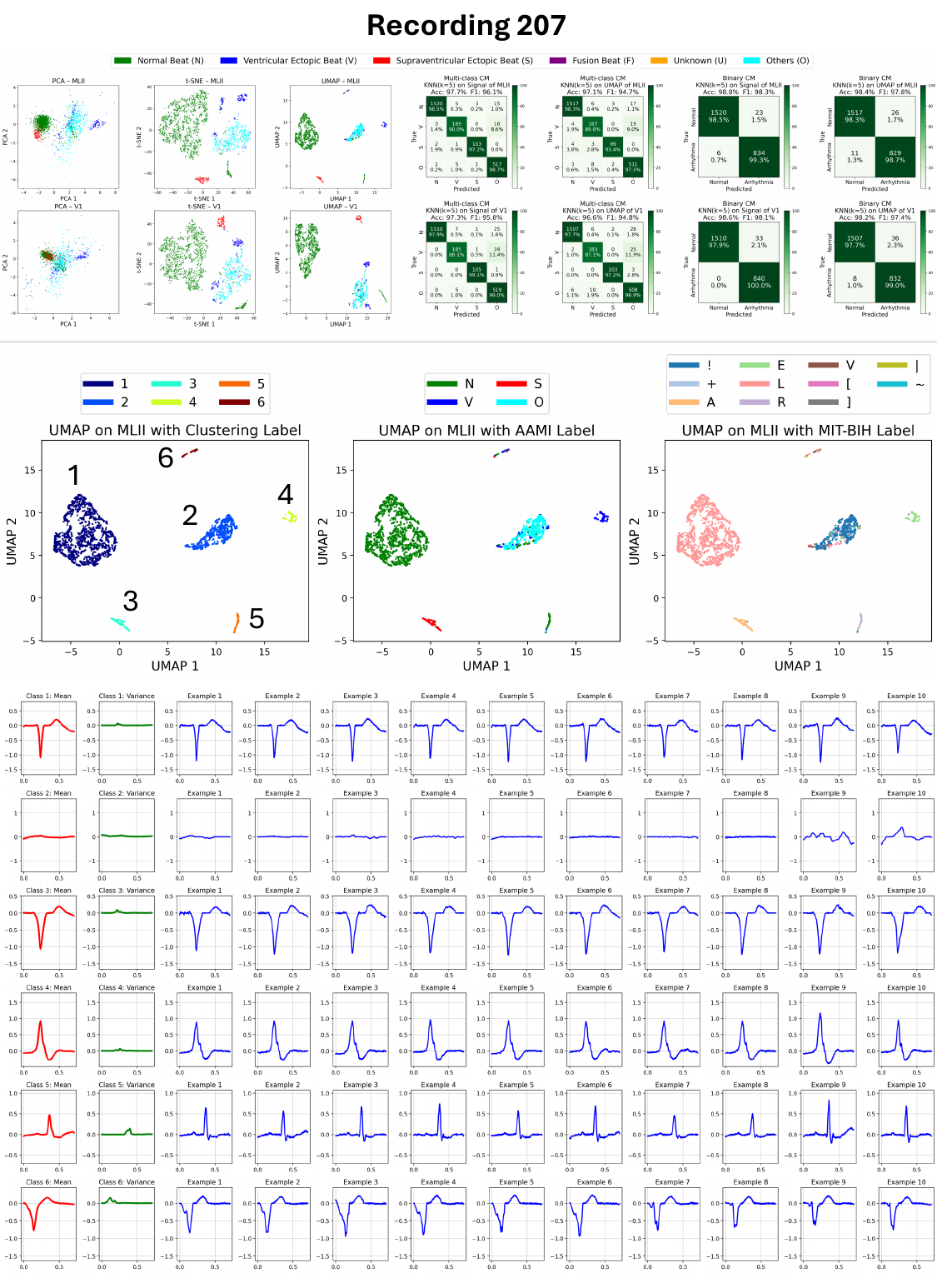}
    \caption{
    Analysis of Recording 207 with dimensionality reduction methods. 
    Top: 2D visualizations using PCA, t-SNE, and UMAP. A k-NN classifier ($k=5$) is applied on the UMAP embeddings \textcolor{black}{and on the original signals} to evaluate the 2D latent space for downstream classification.   
    Bottom: Clustering of 2D UMAP embeddings from the MLII lead, followed by labeling of heartbeats. Ten example signals per cluster are shown, along with their mean and variance. For all AAMI labels, refer to the topmost legend for character definitions. For MIT-BIH labels, see Table~\ref{tab:AAMI_overview}.}
    \label{fig:signal_examples207}
\end{figure*}

\begin{figure*}[htbp]
    \centering
    \includegraphics[width=0.9\textwidth]{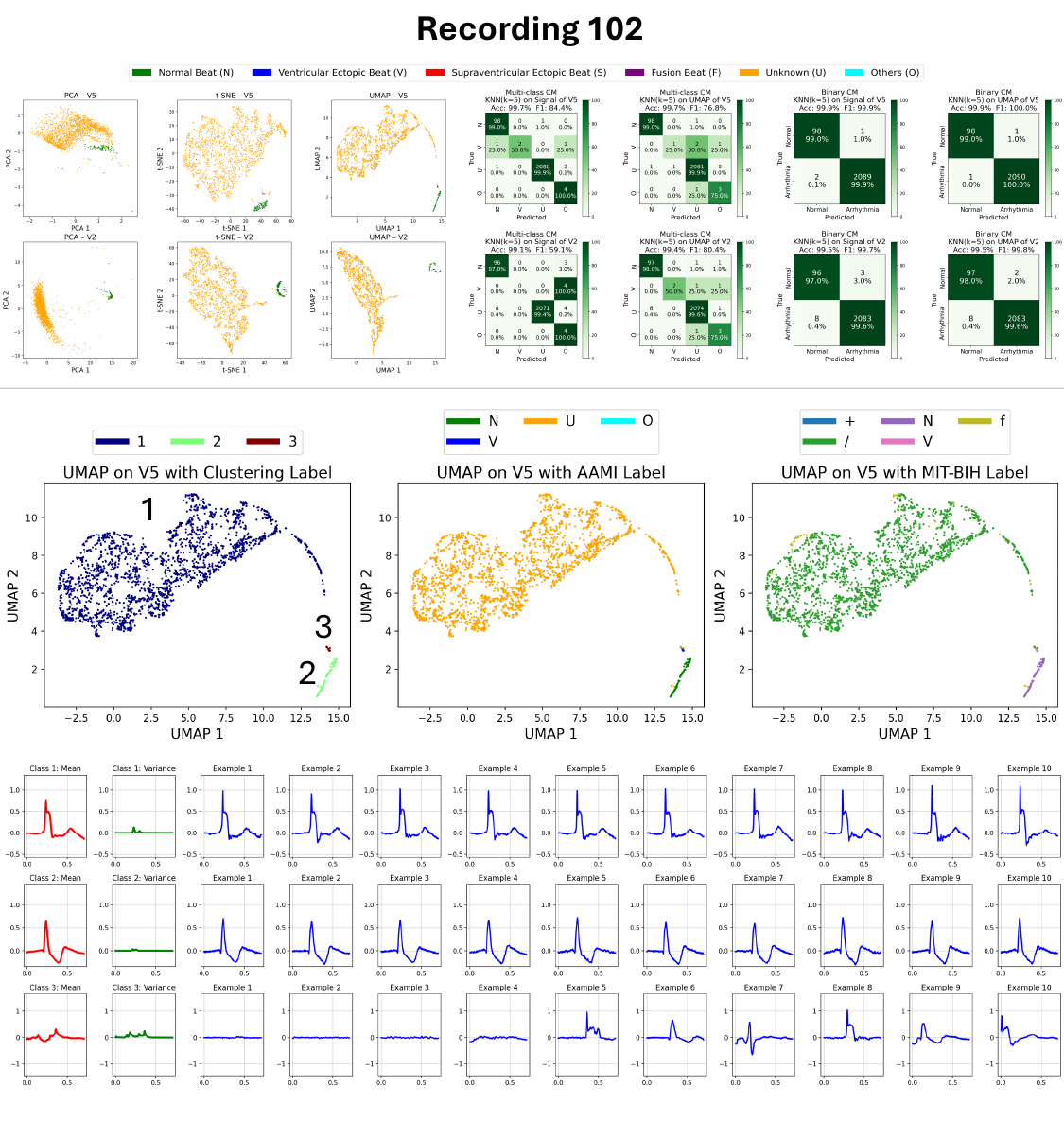}
    \caption{
    Analysis of Recording 102 with dimensionality reduction methods. 
    Top: 2D visualizations using PCA, t-SNE, and UMAP. A k-NN classifier ($k=5$) is applied on the UMAP embeddings \textcolor{black}{and on the original signals} to evaluate the 2D latent space for downstream classification.   
    Bottom: Clustering of 2D UMAP embeddings from the V5 lead, followed by labeling of heartbeats. Ten example signals per cluster are shown, along with their mean and variance. For all AAMI labels, refer to the topmost legend for character definitions. For MIT-BIH labels, see Table~\ref{tab:AAMI_overview}.}
    \label{fig:signal_examples102}
\end{figure*}

\begin{table*}[t]
\centering
\footnotesize
\setlength{\tabcolsep}{2pt}
\caption{\textcolor{black}{Clustering quality metrics and classification performance for different ECG records using UMAP and t-SNE embeddings.}}
\label{tab:clustering_classification_results}
\begin{tabular}{c c c c c c c c c c c}
\toprule
\multirow{3}{*}{Record} & \multirow{3}{*}{Lead} & \multirow{3}{*}{Embedding} 
& \multicolumn{1}{c}{\makecell{Dimensionality\\Reduction\\Metric}} 
& \multicolumn{3}{c}{Clustering Metrics} 
& \multicolumn{2}{c}{\makecell{k-NN (k=5)\\ on Signal}} 
& \multicolumn{2}{c}{\makecell{k-NN (k=5)\\ on 2D Embedding}} \\
\cmidrule(lr){4-4} \cmidrule(lr){5-7} \cmidrule(lr){8-9} \cmidrule(lr){10-11}
 &  & 
 & \makecell{Trustworthiness\\(k=5) $\uparrow$}
 & \makecell{Silhouette\\Coefficient $\uparrow$}
 & \makecell{Davies--Bouldin\\Index $\downarrow$}
 & \makecell{Calinski--Harabasz\\Index $\uparrow$}
 & Accuracy $\uparrow$
 & F1-score $\uparrow$
 & Accuracy $\uparrow$
 & F1-score $\uparrow$ \\
\midrule
\multirow{2}{*}{116} & \multirow{2}{*}{MLII} 
 & UMAP  & 0.98 & \textbf{0.66} & \textbf{0.36} & \textbf{3086.3} & \multirow{2}{*}{99.4} & \multirow{2}{*}{93.3} & \textbf{99.5} & \textbf{94.2} \\
 &  & t-SNE & \textbf{0.99} & 0.36 & 0.62 & 994.3  &  &  & 99.3 & 93.0 \\
\midrule
\multirow{2}{*}{231} & \multirow{2}{*}{V1}
 & UMAP  & 0.97 & \textbf{0.60} & \textbf{0.36} & \textbf{3161.5} & \multirow{2}{*}{99.8} & \multirow{2}{*}{99.4} & 99.7 & 99.2 \\
 &  & t-SNE & \textbf{0.99} & 0.30 & 0.63 & 629.0  &  &  & \textbf{99.8} & \textbf{99.4} \\
\midrule
\multirow{2}{*}{209} & \multirow{2}{*}{MLII}
 & UMAP  & 0.95 & \textbf{0.18} & \textbf{0.61} & \textbf{733.5}  & \multirow{2}{*}{98.7} & \multirow{2}{*}{95.2} & 98.5 & 94.6 \\
 &  & t-SNE & \textbf{0.98} & 0.08 & 0.99 & 489.8  &  &  & \textbf{98.8} & \textbf{95.8} \\
\midrule
\multirow{2}{*}{207} & \multirow{2}{*}{MLII}
 & UMAP  & 0.95 & \textbf{0.72} & \textbf{0.26} & \textbf{5554.7} & \multirow{2}{*}{98.8} & \multirow{2}{*}{98.3} & 98.4 & 97.8 \\
 &  & t-SNE & \textbf{0.98} & 0.37 & 0.53 & 1400.6 &  &  & \textbf{98.7} & \textbf{98.1} \\

 \midrule
\multirow{2}{*}{102} & \multirow{2}{*}{V5}
 & UMAP  & 0.97 & \textbf{0.48} & \textbf{0.48} & \textbf{394.6} & \multirow{2}{*}{99.9} & \multirow{2}{*}{99.9} & 99.9 & 100.0 \\
 &  & t-SNE & \textbf{0.99} & 0.24 & 0.78 & 164.7 &  &  & \textbf{99.9} & \textbf{100.0} \\
\bottomrule
\end{tabular}
\end{table*}

\section{Discussion}

With no labels or prior information, the only starting point for unsupervised NLDR algorithms is a straightforward difference between data points. All three algorithms here, PCA, t-SNE, and UMAP, are based on the covariance matrix, with the former using it directly and the latter two using it as a foundation for distribution functions. As PCA is a linear algorithm that lowers dimension by discarding all but a few (in this case two) directions of maximal variance, it is insufficient for data clustering except in the simplest cases. Both t-SNE and UMAP find nonlinear mappings to the lower-dimensional space, with the former using a single global distribution and the latter using local distributions. Consequently, UMAP better preserved the graph structure of the original data, producing tighter clusters. 

\textcolor{black}{The reliance on statistics means that a population of heartbeats is needed for analysis (even for individual patients). Within the distribution, arrhythmias are sparse and intermittent, and a sufficient number of them must be present to form a distinct cluster in the 2D latent space. Accumulating enough data, and re-running the algorithms to cross-correlate new signals, means that dimensionality reduction is a relatively slow process. On the other hand, if a database is already established, then parametric mappings can be learned (e.g. from a separate neural network) and applied in real time to each new sample \cite{pmlr-v5-maaten09a, sainburg_parametric_2020}.}

\textcolor{black}{Beyond the formation of clusters, the 2D embeddings produced by NLDR can serve as compact feature representations for downstream tasks. Notably, k-NN classification on the 2D embeddings matches or exceeds the performance of k-NN applied to the original high-dimensional signals.  That is, the reduced spaces preserve and often improve the relationship structure between the original data points.}

\textcolor{black}{As NLDR algorithms do not use and cannot provide explicit labels, interpretation of the clusters depends on the user. From a signal processing perspective, sampling points within and between clusters enables comparison of heartbeats based solely on the waveform pattern. Even without medical expertise, different patterns can be discriminated; importantly, some can be identified as erroneous or misclassified directly (e.g. flatlines and partial recordings), providing an automated method of data cleaning \cite{islam_outlier_2025} and iterated improvements.}

\textcolor{black}{With domain knowledge, many of the waveform patterns can be interpreted as medically meaningful. Indeed, this is the traditional form of medical training: correlating signature features with cardiac conditions~\cite{cook_accuracy_2020, breen_ecg_2022}. Here, the task was made easier by focusing on a very limited and well-known dataset. The patients of the MIT-BIH collection had severe (and thus easily identifiable) cardiac conditions. From the viewpoint of bioinformatics, it is a simple system on which to test advanced methods. For broader validation, a wider range of ECG signals is needed \cite{torabi_descriptor_2025}. 
}

\textcolor{black}{The primary contribution of this work is not to surpass supervised classification methods in accuracy, but rather to enable interpretable, patient-specific analysis in an unsupervised manner. A key example is the analysis of multiple leads. While supervised methods must be trained separately for each lead, the bare NLDR methods used here do not have this problem. We applied the same algorithms, without adjustment, to independent leads of the MIT-BIH dataset, and the results were both robust and complementary. For example, they each recognized anomalies that were distinct from conventional arrhythmias, including differences in patient anatomy and electrode placement, that can confuse other types of machine learning algorithms. We also used their different perspectives on the same underlying heartbeat to facilitate analysis, e.g. for Recording 231 in Figure \ref{fig:signal_examples231}. 
}

\textcolor{black}{Leveraging multiple views is naturally more informative than a single lead. In terms of (unsupervised) machine learning, this involves both manifold alignment of the different mappings \cite{islam_manifold-aligned_2022} as well as their projection onto the underlying ECG pathways \cite{man_vectorcardiographic_2015}. Similarly, better understanding will arise from more modalities \cite{baraeinejad_clinical_2023} and the fusion of different sensor measurements ~\cite{wang_fusion_2025, wang_few-shot_2025}, e.g. the confirmation of cardiomyopathies in ECGs using ultrasound \cite{altintepe2026}. For combined analysis, the graph methods discussed here will prove invaluable, as they work for any type of data and can discover new categories without regard for existing labels. Determining the links between these categories and features of physiological or clinical importance is the subject of future research.
}

\section{Conclusion}
We have introduced unsupervised NLDR methods for the analysis of ECG signals. Unlike their supervised counterparts, these algorithms do not require prior knowledge, are not restricted to fixed features, do not require separate training for each lead, and can accommodate any number of variables or characteristics. These qualities make them robust to many issues plaguing biomedical monitoring in general and ECG signals in particular, including missing or repeated data, noise, technician and patient variability (esp. electrode placement) and measurement range. The approach is also well suited for search and discovery in which
features and/or their relationships are unanticipated. We considered three unsupervised dimensional reduction algorithms: PCA, t-SNE, and UMAP. The first is a linear method that severely restricts the types of mappings available; it proved unsuitable for ECG signals. The latter two methods were nonlinear and thus more powerful, but inherently more abstract. Nevertheless, their ability to sort the data into distinct clusters led to a high degree of interpretability. This was achieved by looking at the points (waveforms) within each cluster, which revealed several layers of discrimination. When applied to the total set of heartbeats, nearly all the clusters formed by t-SNE and UMAP corresponded to individual patients. This supports previous observations that ECG signals, like other biometrics, are unique to the individual. In terms of the algorithms, this suggests that the biomarkers present in the cardiac “fingerprints” have stronger correlations than any signatures of arrhythmias. When we applied t-SNE and UMAP to individual patients, the algorithms produced visually separable clusters in 2D, with different clusters corresponding to distinct types of arrhythmias. As with all studies based on the MIT-BIH dataset, our sampling size was small. It was further reduced by our decision to use data from only the MLII and V1 leads, resulting in a total number of 40 patients. Nevertheless, the high performance of the algorithms was promising. Particularly encouraging was the ability of the untrained algorithms to reveal the strengths of each lead independently, viz, the correlation of arrhythmia pathways and timing with the physical geometry of the heart. With more data from more modern ECG systems, including and especially signals from all 12 leads in the current standard of care, unsupervised machine learning has the potential to find new cardiac arrhythmias and to better categorize and act on those that are known.

\section*{Declaration of competing interest}
The authors declare that they have no known competing financial interests or personal relationships that could have appeared to influence the work reported in this paper.

\section*{Code Availability}
All code used in this study is available in the following GitHub repository:
\href{https://github.com/amirrezavazifeh/ECG-with-Manifold-Learning}{https://github.com/amirrezavazifeh/ECG-Heart-Arrhythmia-Detection}.

\section*{Funding}
The authors gratefully acknowledge financial support from the Schmidt DataX Fund at Princeton University made possible through a major gift from the Schmidt Futures Foundation.

\bibliographystyle{unsrt}
\bibliography{References}

\end{document}